\documentclass[11pt]{article}
\usepackage[final]{acl}

\usepackage{times}
\usepackage{latexsym}

\usepackage[T1]{fontenc}
\usepackage[utf8]{inputenc}
\usepackage{microtype}
\usepackage{inconsolata}
\usepackage{graphicx}
\usepackage{booktabs}
\usepackage{colortbl}
\usepackage{xspace}
\usepackage[most]{tcolorbox}
\usepackage{tcolorbox}
\usepackage{array}
\usepackage{amsfonts}
\usepackage{amsmath}

\usepackage[dvipsnames]{xcolor}
\usepackage[table]{xcolor}
\definecolor{lightgray}{gray}{0.95}
\usepackage{amssymb}
\usepackage{multirow}
\usepackage{multicol}
\usepackage{listings}
\usepackage{placeins}
\usepackage{subfigure}
\usepackage{graphicx}
\usepackage{enumitem}
\usepackage{hyperref}
\usepackage{tabularx}
\usepackage{mathrsfs}
\usepackage{pifont}
\usepackage{todonotes}

\hypersetup{
    colorlinks=true,
    linkcolor=purple,
    urlcolor=purple
}

\newcommand{\ours}{\textsc{EvoNote}}
\newcommand{\ds}[1]{\textsc{MM-HealthCN}}
\newcommand{\judge}[1]{\textsc{HealthJudge}}

\definecolor{oursbg}{RGB}{236,255,255}
\newcommand{\ourcell}[1]{\cellcolor{oursbg}#1}

\newtcolorbox{prompt}[1]{colback=gray!5,colframe=gray!35!black,fonttitle=\bfseries, title={#1}}
\newtcolorbox{simple_prompt}{
  colback=gray!5,
  colframe=gray!35!black
}

\usepackage{xcolor, soul,todonotes} 
	\definecolor{kmycolor}{rgb}{0.858, 0.188, 0.478}

\title{Better with Experience: Self-Evolving LLM Agents \\ for Evidence-Grounded Health Community Notes}

\author{
\textbf{Zihang Fu\textsuperscript{1}} \quad
\textbf{Fanxiao Li\textsuperscript{2}} \quad
\textbf{Jianyang Gu\textsuperscript{3}} \quad
\textbf{Haonan Wang\textsuperscript{1}}
\\
\textbf{Preslav Nakov\textsuperscript{4}} \quad
\textbf{Bryan Hooi\textsuperscript{1}} \quad
\textbf{Min-Yen Kan\textsuperscript{1}} \quad
\textbf{Jiaying Wu\textsuperscript{1}\thanks{Corresponding author.}}\\[3pt]
\textsuperscript{1}National University of Singapore, \textsuperscript{2}Yunnan University, \\
\textsuperscript{3}The Ohio State University, 
\textsuperscript{4}Mohamed bin Zayed University of Artificial Intelligence \\[3pt]
\texttt{zihangfu@u.nus.edu, jiayingwu@u.nus.edu}}

\begin{document}
\maketitle
\begin{abstract}
Large Language Model (LLM)-augmented Community Notes offer a scalable path for timely, evidence-grounded correction of health misinformation on social platforms. However, they still reset at every post, leaving useful correction experience from prior cases unused. We introduce \textbf{\ours{}}, an agentic framework that enables health Community Notes generation to \textbf{self-evolve} through an evolving experience memory of prior misinformation correction episodes. Its core is \textbf{fine-grained credit assignment}: \ours{} grounds trajectory-level feedback in health-specific note qualities and distills it into action-level memory for claim analysis, evidence acquisition, and note writing. We evaluate \ours{} on \ds{}, a 1.2K-instance multimodal benchmark of user-flagged health posts with human-written Community Notes and crowd-derived helpfulness labels. Under a human-validated hierarchical utility judge, \ours{}-generated notes are preferred over corresponding human-written notes in \textbf{89.6\%} of cases; on a separate set of \textit{Needs More Ratings} posts without a crowd helpfulness verdict, \ours{} produces helpful notes for \textbf{82.0\%} of cases. It also reduces time to an available candidate correction from multi-hour post-to-note delays to \textbf{under 2 minutes} on average. Analyses link these quality gains to stronger evidence use and reusable correction strategies, positioning self-evolving note generation as a promising paradigm for scalable health misinformation governance.\footnote{Code and data: \url{github.com/Maekami/EvoNote}}
\end{abstract}

\section{Introduction}

\begin{figure}[t]
    \centering
    \includegraphics[width=\columnwidth]{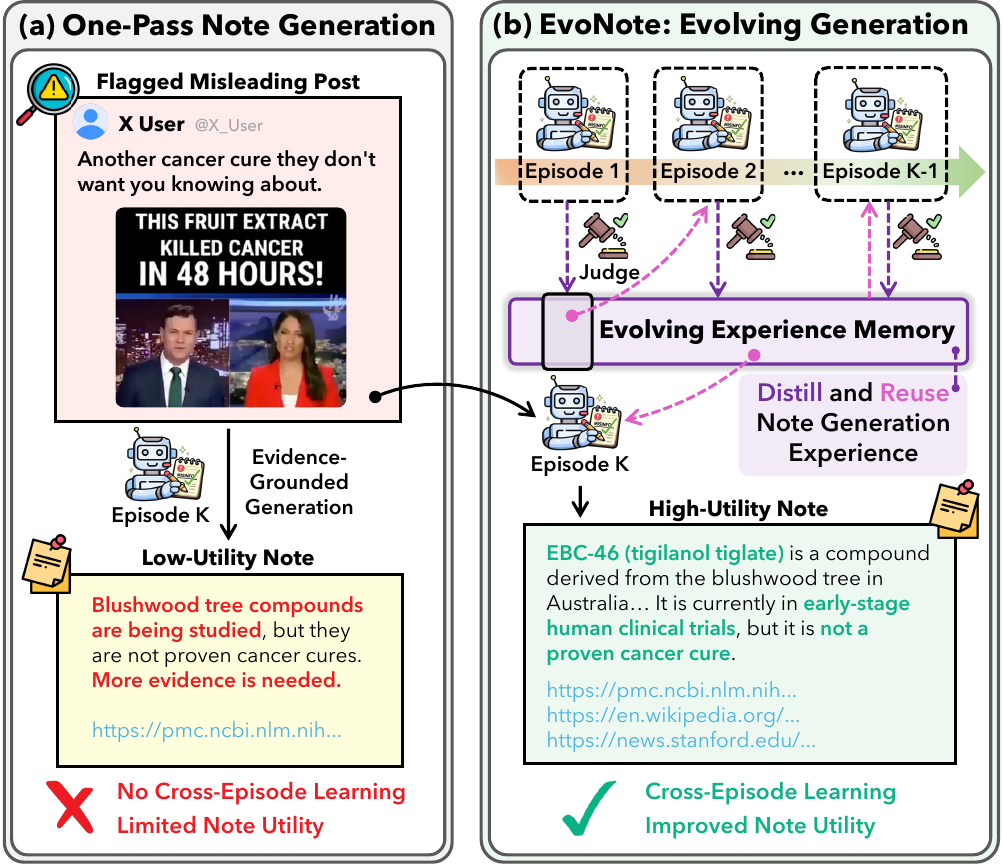}
\caption{
\textbf{Paradigms for LLM-augmented Community Notes generation.}
\textbf{(a)} Existing pipelines treat each episode independently.
\textbf{(b)} \textsc{EvoNote} distills prior episodes into evolving memory to improve future notes.
}
\label{fig:intro}
\vspace{-1em}
\end{figure}

Health misinformation on social platforms requires timely, evidence-backed, and user-understandable correction \cite{vraga2020correction,ecker2022psychological,kbaier2024prevalence}. Community Notes \cite{wojcik2022birdwatch} on X offers a scalable and increasingly influential governance paradigm for this task.

It turns crowd-sourced verification and community ratings into concise, source-grounded corrections that can be shown alongside misleading posts \cite{renault2024collaboratively,slaughter2025community}.

\textbf{LLM-augmented Community Notes} \cite{wu2026beyond,singh2026gitsearch} further improve the timeliness and helpfulness of correction through automated, evidence-grounded note generation. However, as shown in Figure~\ref{fig:intro}, current pipelines still follow a \textbf{one-pass} paradigm, treating each post as an isolated instance. This formulation fits poorly with real-world health misinformation governance, where misleading claims arrive as a chronological stream and often recur through similar rhetorical strategies \cite{shaar-etal-2020-known,ijcai2021p619,skafle2022misinformation} and evidence needs \cite{schneider2021effects}. Our analysis of 4.9K health-related notes published over 41 months (\S\ref{sec:analysis}) confirms this recurrence and shows that it creates reusable correction experience: earlier helpful notes often reveal transferable evidence-seeking and writing strategies for later posts of the same type. 

However, under a one-pass formulation, this experience remains episode-local, leaving later corrections vulnerable to repeated, avoidable errors. This motivates \textbf{self-evolving note generation}: systems that turn prior note-generation episodes into reusable experience to guide future notes.

Motivated by this, we introduce \textbf{\ours{}}, an agentic framework that enables self-evolving health Community Notes generation through an \textbf{evolving experience memory} of prior misinformation correction episodes.
For each note-generation trajectory, which spans claim analysis, evidence acquisition, and note writing, a \textbf{Social Utility Judge} evaluates the trajectory using four health-specific note qualities grounded in prior work on health communication: \textbf{(1)} understandability \cite{sorensen2012health}, \textbf{(2)} meaningfulness for health interpretation and risk perception \cite{ferrer2015risk}, \textbf{(3)} usability for safe public action \cite{denniss2022development}, and \textbf{(4)} trustworthiness under evidence uncertainty \cite{kington2021identifying}. 
A \textbf{Memory Evolver} then distills these judgments into fine-grained strategies, assigning each lesson to the phase where it can guide future generation.
At inference time, \ours{} retrieves relevant memories to guide claim analysis, budget-aware web evidence acquisition, and evidence-grounded note writing without updating model parameters.

To evaluate \ours{}, we curate \textbf{\ds{}}, a 1.2K-instance multimodal benchmark of flagged health posts paired with human-written Community Notes, crowd-derived helpfulness labels, and a human-validated protocol for hierarchical note utility evaluation (\S\ref{sec:benchmark}). On \ds{}, offline pairwise evaluation (\S\ref{sec:main-results}) shows that \ours{}-generated notes are preferred over human-written Community Notes in \textbf{89.6\%} of cases under a human-validated utility judge, while also outperforming LLM-augmented Community Notes baselines. On unresolved \textit{Needs More Ratings} cases before crowd consensus, \ours{} produces helpful notes for \textbf{82.0\%} of cases. It also reduces time to an available candidate correction from multi-hour post-to-note delays to \textbf{under 2 minutes} on average. Further analyses (\S\ref{sec:discussion}) link these quality gains to more diverse and higher-quality evidence sources and reusable memories that encode recurring correction strategies. These results point to a central design principle for health misinformation governance: every flagged post should become experience before the next similar one recurs.

\section{Related Work}
\label{sec:related-work}

\paragraph{Community Notes and LLM Augmentation.}
Crowd-sourced fact-checking provides a scalable response to online misinformation \cite{allen2021scaling,martel2024crowds,shahbazi2024social,pfander2025spotting,xing2026communitynotes}, and Community Notes on X has been shown to reduce misinformation engagement \cite{slaughter2025community} and support balanced discourse \cite{chuai2024community}.
Its impact, however, depends on useful notes being written, rated, and surfaced in time \cite{renault2024collaboratively,wu2026beyond}, creating a bottleneck for health misinformation, where rapid early spread makes correction delay especially costly \cite{kouzy2020coronavirus,van2022misinformation}.
This bottleneck has motivated LLM augmentation for note creation, either by aggregating multiple existing \textit{Needs More Ratings} notes \cite{de2025supernotes} or by generating one-pass evidence-grounded notes for individual posts \cite{wu2026beyond,singh2026gitsearch}.
The first path depends on multiple human candidates for the same post; the second resets after each post.
\ours{} fills this gap by turning each agentic correction trajectory into reusable action-level memory for future note generation.

\paragraph{Memory-Augmented Agents.}
Prior work equips LLM agents with experience reuse through passive memory over retrieved documents, dialogue histories, or stored facts \cite{lewis2020retrieval,packer2024memgpt,zhong2024memorybank,xie-etal-2025-fire}; managed long-term memory with add, merge, delete, and retrieval operations \cite{chhikara2025mem0,rasmussen2025zep,li2025memos}; and trajectory-based memory that stores reflections, procedural hints, or reusable plans from prior interactions \cite{shinn2023reflexion,zhao2024expel,suzgun2026dynamic}.
Recent self-evolving memory work, such as Evo-Memory \cite{wei2025evo}, studies experience reuse across continuous task streams.
\ours{} builds on this direction but goes beyond retrieving past experience by assigning it to the correction stage where it should guide action.
For health Community Notes generation, this means distilling trajectory-level feedback into health-grounded, phase-specific strategies for claim analysis, evidence acquisition, and note writing.
This positions \ours{} as memory-level self-evolution under a fixed backbone and workflow, complementary to broader methods that update prompts, workflows, or policies \cite{wang2025ragen,he2026evotest}.

\begin{figure*}[ht!]
    \centering
    \includegraphics[width=\textwidth]{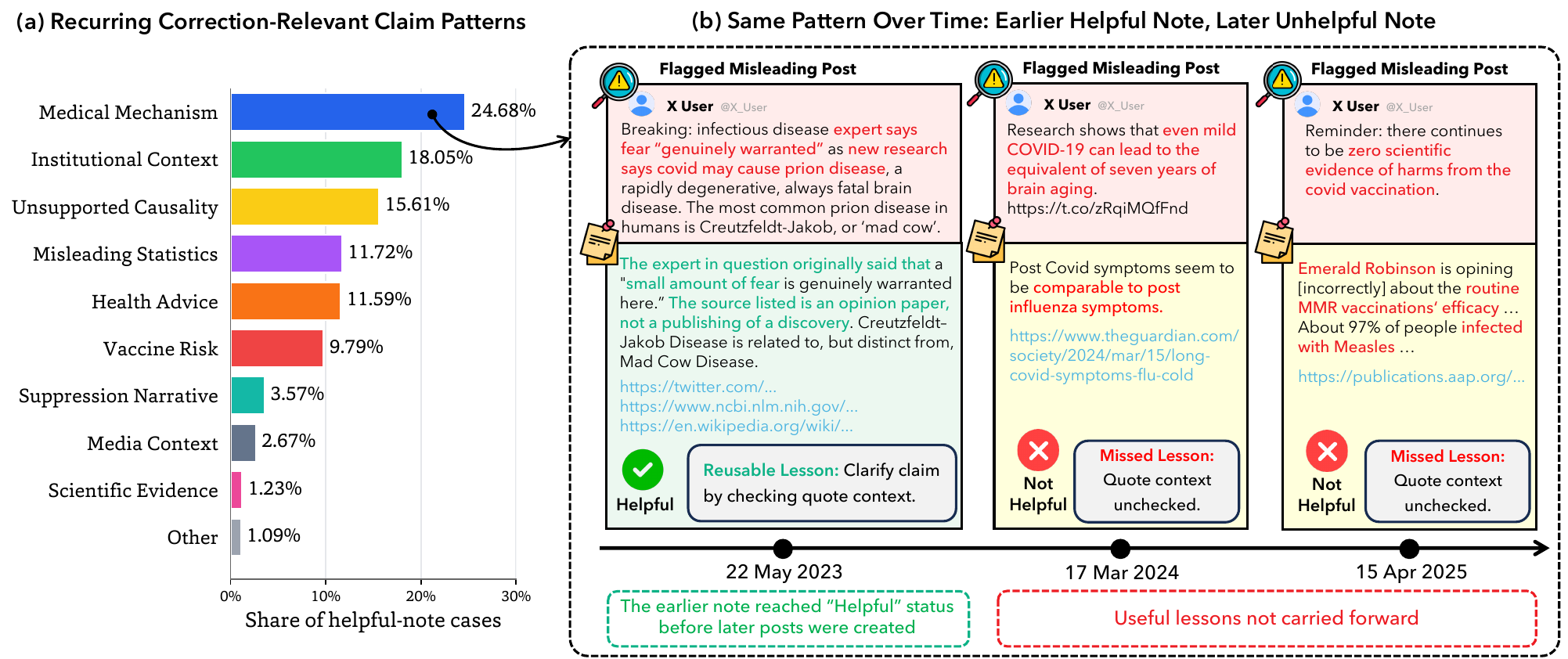}
\caption{
\textbf{Reusable misinformation correction experience can remain episode-local.}
\textbf{(a)} Distribution of major correction-relevant claim patterns inferred from helpful-note cases; detailed categories are provided in Table~\ref{tab:claim-patterns}.
\textbf{(b)} Timeline of one recurring medical mechanism overclaim pattern. Although the earlier \textit{Helpful} lesson was available before the later posts appeared, later notes miss the same correction strategy and are rated \textit{Not Helpful}.
}
\label{fig:recur-analysis}
\end{figure*}

\section{Motivating Analysis: Reusable Misinformation Correction Experience}
\label{sec:analysis}

Health misinformation correction on social platforms unfolds as a stream, and misleading mechanisms and evidence needs often recur over time \cite{skafle2022misinformation}. 
This makes earlier helpful notes a potential source of reusable correction experience. 
We analyze health Community Notes on X to ask two questions: \textbf{(1)} Do misleading health posts recur through correction-relevant claim patterns? \textbf{(2)} Can later notes still fail on a pattern that has already received a helpful correction?

\subsection{Data and Claim Pattern Extraction}
\label{sec:analysis-setup}

\noindent\textbf{Data Scope.} 
We collect publicly available, user-contributed Community Notes on X\footnote{\url{https://x.com/i/communitynotes/download-data}} up to 14 October 2025 and retain English entries for consistency. Following prior work \cite{wu2026beyond}, we identify health-related notes through topic-oriented filtering with Lingshu-32B \cite{xu2025lingshu}, a medical-specific LLM, and retrieve the corresponding flagged misleading posts from X. This yields 48,295 notes addressing 40,116 health-related flagged posts. Among them, 6,336 (13.12\%) notes have crowd-derived helpfulness labels (\textit{``Helpful''}: 4,874; \textit{``Not Helpful''}: 1,462); the remaining cases are unresolved, i.e., \textit{``Needs More Ratings''}. Full data setup is provided in Appendix~\ref{app:analysis-data-filter}; additional latency and visibility statistics are reported in Appendix~\ref{app:latency-visibility}. 

\noindent\textbf{Claim Pattern Extraction.} 
We derive correction-relevant claim patterns from 4,874 helpful human-written notes published between May 2022 and October 2025. Helpful notes provide high-precision signals for misleadingness attribution by revealing what aspect of the post required correction and what evidence or framing made the correction useful. Two human experts collaboratively reviewed 200 sampled note-post pairs to summarize the main correction causes and to construct a compact taxonomy of misleading mechanisms (Table~\ref{tab:claim-patterns}). We then use GPT-4.1 \cite{openai2025gpt4_1} in order to classify the remaining helpful-note cases under this taxonomy. For posts involving multiple mechanisms, we assign the primary pattern according to which correction strategy is most central to the helpful note. Detailed category definitions are provided in Appendix~\ref{app:analysis-claim-pattern}.

\begin{figure*}[t]
    \centering
    \includegraphics[width=\textwidth]{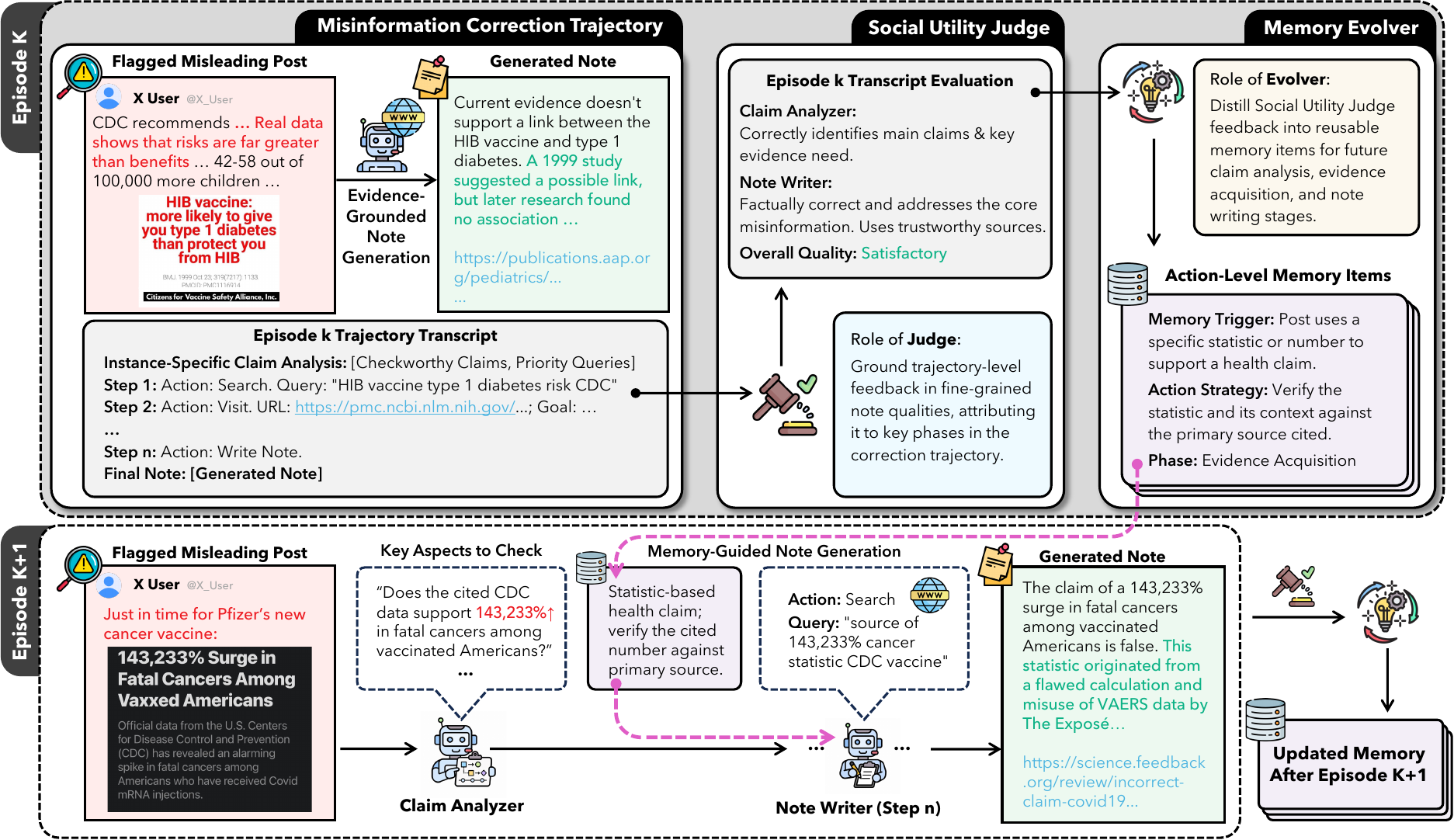}
\caption{
\textbf{Overview of \ours{} for self-evolving health Community Notes generation.} \ours{} turns trajectory-level feedback into evolving experience memory to inform future note-generation episodes.}
\label{fig:overview}
\end{figure*}

\subsection{Useful Lessons Remain Episode-Local}
\label{sec:analysis-findings}

Our analysis reveals two key observations. \textbf{(1) Recurring claim patterns create reusable correction needs.} As shown in Figure~\ref{fig:recur-analysis}(a), the cases cluster around correction-relevant patterns, such as medical mechanism misrepresentations and unsupported causality. 
These patterns imply transferable strategies, such as verifying source context, separating association from causation, and decomposing biomedical mechanism overclaims. A finer-grained chronological analysis (Appendix \ref{app:logic-recurrence}) further shows substantial cross-post similarity in correction logic. 

\textbf{(2) Useful lessons can remain episode-local.} 
Figure~\ref{fig:recur-analysis}(b) shows a recurring medical mechanism overclaim where an earlier \textit{Helpful} note reveals a reusable lesson: clarify the claim by checking its quote context. The lesson was already available before the later posts appeared. Subsequent notes for the same correction-relevant pattern still leave the quote context unchecked and are rated \textit{Not Helpful}. This case shows that useful correction experience can remain local to the episode that produced it and fail to carry forward to later similar cases.

\section{\ours{}: Self-Evolving Health Community Notes Generation}
\label{sec:method}

The analysis in Section~\ref{sec:analysis} points to a concrete opportunity: \textit{recurring misinformation patterns expose lessons for later corrections}. Motivated by this, \ours{} turns each correction episode into action-level memory for future note generation. As shown in Figure~\ref{fig:overview}, a Social Utility Judge grounds trajectory-level feedback in health-specific note qualities, and a Memory Evolver distills it into phase-specific strategies for claim analysis, evidence acquisition, and writing. By learning from its own trajectories, \ours{} adapts to emerging claims even when reliable human notes are sparse, delayed, or unavailable. Framework details and a full case are provided in Appendices~\ref{app:framework} and~\ref{app:demo}.

\subsection{Note-Generation Episode Formulation}
\label{sec:notegen-episode}

Let $x_k$ denote the flagged post in episode $k$, which may contain text, image, or video content. We convert $x_k$ into a unified textual representation $\tilde{x}_k$ through captioning, preserving the multimodal information needed to identify and verify the health claim. This allows text, image, and video cases to share the same agentic correction trajectory and supports medical-capable LLMs with strong tool-use and multi-step reasoning abilities.

A note-generation episode produces a trajectory
\[
\tau_k = (\tilde{x}_k, c_k, H_k, E_k, y_k),
\]
where $c_k$ is the instance-specific claim analysis, $H_k$ is the action history (e.g., ``search, visit, ...''), $E_k$ is the collected evidence, and $y_k$ is the final generated note.
The claim analysis identifies the core check-worthy claims, priority evidence needs, and intended correction focus for the current episode.

\subsection{Trajectory-Level Utility Judgment}
\label{sec:social-utility-judge}

At the core of \ours{} is an \textbf{evolving experience memory} that turns completed correction episodes into actionable guidance. A final note reveals correction quality, while its trajectory exposes the decisions behind that quality. \ours{} converts this trajectory-level signal into structured feedback through a \textbf{Social Utility Judge} agent.

Once the full trajectory $\tau_k$ for episode $k$ is obtained, the Judge evaluates it and produces fine-grained feedback $f_k$.
The Judge grounds feedback in a health-correction utility framework drawn from prior health communication desiderata (\S\ref{app:social-utility-judge}): \textbf{(1) understandability} for accessible comprehension \cite{sorensen2012health,warde2018plain}, \textbf{(2) meaningfulness} for health interpretation and risk perception \cite{ferrer2015risk,prike2023effective}, \textbf{(3) usability} for safe action implications \cite{heydari2021effect,denniss2022development}, and \textbf{(4) trustworthiness} for calibrated evidence use under uncertainty \cite{schneider2021effects,denniss2022development}.
This framework captures how a correction becomes useful to the public, moving from comprehension, to interpretation, to action, to calibrated trust. The resulting feedback maps note utility back to trajectory decisions, revealing what the agent should preserve, revise, or avoid in future claim analysis, evidence acquisition, and writing.

Based on $f_k$, the judge further assigns a utility mode $u_k$: \textbf{(1)} \textsc{Unsatisfactory}, \textbf{(2)} \textsc{Satisfactory}, or \textbf{(3)} \textsc{Excellent}. This mode acts as a routing signal for memory construction: \textsc{Unsatisfactory} routes the episode toward failure-avoidance memories, \textsc{Satisfactory} toward improvement-oriented memories, and \textsc{Excellent} toward success-reuse memories.
Thus, $f_k$ diagnoses \textit{what} should be learned from the trajectory, while $u_k$ summarizes \textit{how} the episode should update memory.

\subsection{Actionable Memory Evolution}
\label{sec:memory-evolver}

Given the memory-update mode $u_k$, the key challenge is to make the completed episode useful beyond the current case.
A reusable memory must specify when it applies, what strategy it recommends, and which note-generation phase it should guide. The \textbf{Memory Evolver} converts the completed trajectory into such action-level memory as follows:
\[
\begin{aligned}
\Delta \mathcal{M}^{k} &= \textsc{Distill}(\tau_k, u_k), \\
\mathcal{M}^{k} &= \mathcal{M}^{k-1} \cup \Delta \mathcal{M}^{k}.
\end{aligned}
\]
The mode $u_k$ determines whether the distilled items emphasize failure avoidance, targeted improvement, or success reuse.

Each memory item in $\Delta \mathcal{M}^{k}$ is represented as follows:
\[
m = (\textsc{Trigger}, \textsc{ActionStrategy}, \textsc{Phase}).
\]
\textsc{Trigger} specifies when the memory should be retrieved, \textsc{ActionStrategy} states the reusable correction strategy, and \textsc{Phase} indicates where it applies: claim analysis, evidence acquisition, or note writing.
This structure turns episode-level feedback into operational guidance for future episodes.
For example, if a trajectory mishandles a statistic-based health claim, the Evolver can create an evidence-acquisition memory that triggers on similar numerical claims and recommends checking the original source, denominator, and context before writing.

\subsection{Memory-Guided Note Generation}
\label{sec:memory-guided-generation}

At episode $k+1$, \ours{} uses the current memory $\mathcal{M}^{k}$ to guide a new note-generation trajectory. 
Given the textual post representation $\tilde{x}_{k+1}$, generation proceeds in two stages: memory-guided claim analysis, followed by budgeted evidence acquisition and note writing.

\paragraph{Claim Analysis.}
The \textbf{Claim Analyzer} agent first summarizes the post into a compact retrieval query $q_C(\tilde{x}_{k+1})$ that captures the correction-relevant claim pattern. 
It retrieves phase-matched memories from $\mathcal{M}^{k}$ by selecting top-$n$ items whose \textsc{Phase} is claim analysis and whose \textsc{Trigger} has the highest lexical overlap with the query:
\[
\begin{aligned}
R_C^{k+1}
&= \textsc{Retrieve}\bigl(
\mathcal{M}^{k}, q_C(\tilde{x}_{k+1}), \\
&\qquad
\textsc{Phase}=\textsc{ClaimAnalysis}
\bigr).
\end{aligned}
\]
The retrieved memories return their \textsc{ActionStrategy} fields, which guide the analyzer in producing an instance-specific verification brief $c_{k+1}$:
\[
c_{k+1} = C(\tilde{x}_{k+1}, R_C^{k+1}),
\]
which specifies the check-worthy claims and priority verification questions that the subsequent note-generation trajectory should address.

\paragraph{Budgeted Evidence Acquisition and Note Writing.}
Given $c_{k+1}$ and the post representation $\tilde{x}_{k+1}$, the \textbf{Note Writer} agent performs a budgeted action trajectory indexed by step $t$.
Before each action step, it summarizes the current decision state into a retrieval query $q_A^t$, capturing the search history, collected evidence, remaining uncertainties, and current phase $p_t$, where $p_t \in \{\textsc{EvidenceAcquisition}, \textsc{NoteWriting}\}$.
The writer first uses $p_t$ to select the phase-matched memory subset, then retrieves top-$n$ memories by lexical overlap between $q_A^t$ and memory triggers:
\[
\begin{aligned}
R_A^{k+1,t}
&= \textsc{Retrieve}\bigl(
\mathcal{M}^{k},
q_A^t,
\textsc{Phase}=p_t
\bigr).
\end{aligned}
\]
The retrieved \textsc{ActionStrategy} fields guide the next action.
The Writer follows separate budgets for \textsc{Search} and \textsc{Visit}, denoted by $B_S$ and $B_V$.
The trajectory starts with evidence-seeking actions:
\[
\mathcal{A}_t = \{\textsc{Search}, \textsc{Visit}\}, 
\quad a_t \in \mathcal{A}_t.
\]
\textsc{Search} issues a web query, while \textsc{Visit} inspects either a URL returned by search or a URL contained in the original post, then extracts correction-relevant evidence.

After at least one search and one visit, $\mathcal{A}_t$ expands to include \textsc{Write} and \textsc{Abstain}:
\[
\mathcal{A}_t \leftarrow \mathcal{A}_t \cup \{\textsc{Write}, \textsc{Abstain}\},
\quad a_t \in \mathcal{A}_t.
\]
At all steps, $\mathcal{A}_t$ is \textbf{budget-aware}: \textsc{Search} is removed once its search budget $B_S$ is exhausted, and \textsc{Visit} is removed once its visit budget $B_V$ is exhausted.
\textsc{Write} drafts a note when evidence is sufficient and optionally refines it for length compliance; \textsc{Abstain} terminates generation when evidence cannot support a reliable correction.

\subsection{Cross-Episode Self-Evolution}
\label{sec:self-evolution}

The completed note-generation trajectory of episode $k+1$ becomes the next source of experience.
It is evaluated by the Social Utility Judge and distilled by the Memory Evolver, updating $\mathcal{M}^{k}$ into $\mathcal{M}^{k+1}$.
Thus, \ours{} forms a closed loop:
\[
\tau_k \rightarrow u_k \rightarrow \Delta \mathcal{M}^{k} \rightarrow \tau_{k+1}.
\]
This loop makes correction experience cumulative without updating model parameters or relying on a fixed bank of human notes, allowing later episodes to reuse strategies distilled from earlier trajectories.

\section{The \ds{} Benchmark}
\label{sec:benchmark}

We introduce \ds{}, a multimodal benchmark for evaluating health Community Notes generation under both binary helpfulness judgment and pairwise social utility comparison.

\paragraph{Data.}
\ds{} contains 1.2K user-flagged, health-related posts on X, each paired with a human-written Community Note and its crowd-derived helpfulness label, either \textit{Helpful} or \textit{Not Helpful}. We construct three modality subsets based on post content: \textsc{Text}, \textsc{Image}, and \textsc{Video}, each with 400 instances balanced between \textit{Helpful} and \textit{Not Helpful} notes. Details of data construction are provided in Appendix~\ref{app:benchmark-data}.

For image and video posts, we use captions as the unified textual input representation. A pilot test with three independent annotators shows that captions reliably preserve correction-relevant content, with majority-vote accuracy of 0.95 on 100 sampled image captions and 0.96 on 100 sampled video captions. Appendices~\ref{app:multimodal-captioning} and \ref{app:caption-validation} detail the captioning process and validation; Appendix~\ref{app:vlm-failure} further supports the necessity of captioning by showing the limited reliability of direct multimodal-VLM instantiation in this agentic health setting.

\paragraph{Hierarchical Utility Judgment.}
We extend the hierarchical evaluation scheme from prior work \cite{wu2026beyond}, which was validated with near-perfect human alignment. We first apply a rule-based \textbf{platform-compliance filter}, marking notes over the 280-character Community Notes limit as \textit{Not Helpful}. GPT-4.1-based judges then assess the remaining notes through \textbf{three progressive gates}: \textbf{(1)} evidence relevance (R), \textbf{(2)} correctness of evidence representation (C), and \textbf{(3)} helpfulness of note style (H). A note is deemed \textit{Helpful} only if it passes the length filter and all three gates.

To compare notes beyond binary helpfulness, we introduce a GPT-4.1-based \textbf{pairwise utility judge} (Appendix \ref{app:pairwise-utility}). When two notes receive the same binary outcome, the judge performs a randomized win/tie/loss comparison on understandability, meaningfulness, usability, and trustworthiness. In a rationale-aware audit, 98\% of 100 randomly sampled comparisons are judged accurate by humans. A blind evaluation shows 91\% agreement with human-majority overall preferences and 90--93\% agreement across the four utility dimensions. Detailed setup, validation protocols, and judge reliability analysis are provided in Appendix~\ref{app:eval-details}.

\begin{table*}[t]
\centering
\resizebox{\textwidth}{!}{%
\begin{tabular}{l
                cccc|
                cccc|
                cccc|
                cc}
\toprule
\multirow{2}{*}{\textbf{Model}} &
\multicolumn{4}{c|}{\textbf{\textsc{Text}}} &
\multicolumn{4}{c|}{\textbf{\textsc{Image}}} &
\multicolumn{4}{c|}{\textbf{\textsc{Video}}} &
\multicolumn{2}{c}{\textbf{Overall}} \\   
\cmidrule(lr){2-5} \cmidrule(lr){6-9} \cmidrule(lr){10-13} \cmidrule(lr){14-15}
& \textbf{R} & \textbf{C} & \textbf{H} & \textbf{Win.}
& \textbf{R} & \textbf{C} & \textbf{H} & \textbf{Win.}
& \textbf{R} & \textbf{C} & \textbf{H} & \textbf{Win.}
& \textbf{H} & \textbf{Win.} \\   
\midrule
Human Baseline
& 73.00 & 53.50 & 35.75 & \ourcell{---}
& 74.75 & 55.50 & 40.00 & \ourcell{---}
& 76.75 & 54.00 & 39.25 & \ourcell{---}
& 38.33 & \ourcell{---} \\
\midrule
GPT-4.1
& 39.75 & 31.50 & 28.50 & \ourcell{69.50}
& 36.25 & 28.25 & 27.00 & \ourcell{63.25}
& 32.75 & 27.50 & 25.25 & \ourcell{65.13}
& 26.92 & \ourcell{65.96} \\
Grok-4.3
& 94.75 & 79.50 & 69.00 & \ourcell{76.50}
& \underline{94.50} & 76.25 & 71.75 & \ourcell{77.88}
& 91.00 & 71.50 & 67.00 & \ourcell{76.38}
& 69.25 & \ourcell{76.92} \\
Gemini-2.5-Flash
& 63.25 & 57.75 & 50.75 & \ourcell{76.13}
& 63.50 & 57.25 & 54.00 & \ourcell{75.88}
& 58.00 & 47.75 & 46.00 & \ourcell{72.88}
& 50.25 & \ourcell{74.96} \\
\midrule
CrowdNotes+
& 88.50 & \underline{83.50} & 68.50 & \ourcell{69.25}
& 87.50 & 76.50 & 65.75 & \ourcell{70.63}
& 87.75 & 71.00 & 60.50 & \ourcell{67.75}
& 64.92 & \ourcell{69.21} \\
DeepResearch
& 95.50 & 81.25 & \underline{73.75} & \ourcell{81.13}
& 92.50 & 75.75 & 70.00 & \ourcell{76.75}
& 90.00 & 75.75 & 73.25 & \ourcell{76.38}
& 72.33 & \ourcell{78.09} \\
ExpRAG
& 83.25 & 66.75 & 59.00 & \ourcell{76.88}
& 91.00 & 76.50 & 70.25 & \ourcell{81.25}
& 89.50 & 74.00 & 68.00 & \ourcell{80.75}
& 65.75 & \ourcell{79.63} \\
ReMem
& 90.25 & 77.50 & 67.00 & \ourcell{76.75}
& 94.00 & 82.00 & 72.00 & \ourcell{72.75}
& \underline{95.25} & 81.00 & 70.25 & \ourcell{74.00}
& 69.75 & \ourcell{74.50} \\
\midrule
\ours{}\textsc{-fm}
& \underline{97.25} & 80.25 & 70.25 & \ourcell{\underline{85.38}}
& 96.50 & \underline{85.25} & \underline{78.25} & \ourcell{\underline{86.13}}
& 88.50 & 78.25 & 71.00 & \ourcell{82.75}
& \underline{73.17} & \ourcell{\underline{84.75}} \\
\ours{}\textsc{-nm}
& 95.50 & 80.50 & 68.50 & \ourcell{85.25}
& 93.00 & 77.25 & 71.00 & \ourcell{82.00}
& 97.00 & \underline{82.00} & \underline{76.00} & \ourcell{\underline{85.50}}
& 71.83 & \ourcell{84.25} \\
\ours{}\textsc{-nm-na}
& 90.75 & 74.25 & 64.50 & \ourcell{81.63}
& 90.50 & 74.50 & 69.00 & \ourcell{79.00}
& 93.50 & 73.50 & 67.50 & \ourcell{77.63}
& 67.00 & \ourcell{79.42} \\
\midrule
\ours{}
& \textbf{97.50} & \textbf{89.25} & \textbf{78.75} & \ourcell{\textbf{89.38}}
& \textbf{96.50} & \textbf{86.75} & \textbf{80.50} & \ourcell{\textbf{89.25}}
& \textbf{97.00} & \textbf{90.50} & \textbf{84.25} & \ourcell{\textbf{90.13}}
& \textbf{81.17} & \ourcell{\textbf{89.59}} \\

\bottomrule
\end{tabular}}
\caption{
\textbf{Effectiveness (\%) of note generation on \ds{}.}
``Human Baseline'' is the original human-written Community Notes.
Metrics are relevance after length-validity filtering (\textbf{R}), evidence correctness (\textbf{C}), note helpfulness (\textbf{H}), and pairwise win rate against the Human Baseline (\textbf{Win.}; ties count as 0.5).
See \S\ref{sec:exp-setup} for setup, \S\ref{sec:main-results} for variants, and Table~\ref{tab:main-results-len} for length compliance rates.
Best and second-best results are \textbf{bolded} and \underline{underlined}.
}
\label{tab:main-results}
\end{table*}

\section{Experiments}

\subsection{Experimental Setup}
\label{sec:exp-setup}

\textbf{Baselines.} We compare \ours{} against human-written Community Notes and three families of automated baselines: \textbf{(1)} web-search-enabled LLMs, including GPT-4.1 \cite{openai2025gpt4_1}, Grok-4.3 \cite{xai2026grok43}, and Gemini-2.5-Flash \cite{comanici2025gemini}, \textbf{(2)} LLM-based Community Notes generation systems, including CrowdNotes+ \cite{wu2026beyond} and DeepResearch-30B-A3B \cite{team2025tongyi}, and \textbf{(3)} memory-augmented LLM agents from Evo-Memory suite \cite{wei2025evo}, including ExpRAG and ReMem. More details are provided in Appendix~\ref{app:baselines}.

\paragraph{Implementation Details.} 
\ours{} uses the medical-specific MedGemma-27B \cite{sellergren2025medgemma} as the backbone LLM. For fair comparison, applicable methods share the same web-search interface, CrowdNotes+, ExpRAG, and ReMem use the same base LLM as \ours{}, and all memory-based methods retrieve top-$k$ memories. 
The full implementation details and all leakage-control constraints are provided in Appendices~\ref{app:implementation-details} and~\ref{app:eval-constraints}, respectively.

\subsection{Main Results: Effectiveness of \ours{}}
\label{sec:main-results}

\paragraph{\ours{} consistently improves correction quality across modalities.}
On the consensus-resolved \ds{} benchmark, \ours{} achieves an average \textbf{89.6\%} pairwise win rate over human-written Community Notes and consistently outperforms all baselines across text, image, and video posts (Table~\ref{tab:main-results}). The below-50\% helpfulness of human-written notes under our hierarchical judge is consistent with prior findings~\cite{wu2026beyond} that crowd-rated helpfulness can mask subtle evidence-related failures, such as weak source relevance or incorrect evidence representation; here, we show that this gap extends across modalities. 

\paragraph{\ours{} rescues failed human notes and strengthens already helpful ones.}
For judge-rated \textit{Not Helpful} human notes in Table \ref{tab:main-results} ($N=740$), failures are attributed to evidence relevance (302 cases; 40.8\%), evidence representation correctness (246; 33.2\%), and note-style helpfulness (192; 25.9\%); \ours{} produces a helpful alternative in \textbf{77.2\%} of these cases.

\begin{figure}[t]
\begin{center}
    \includegraphics[width=\linewidth]  {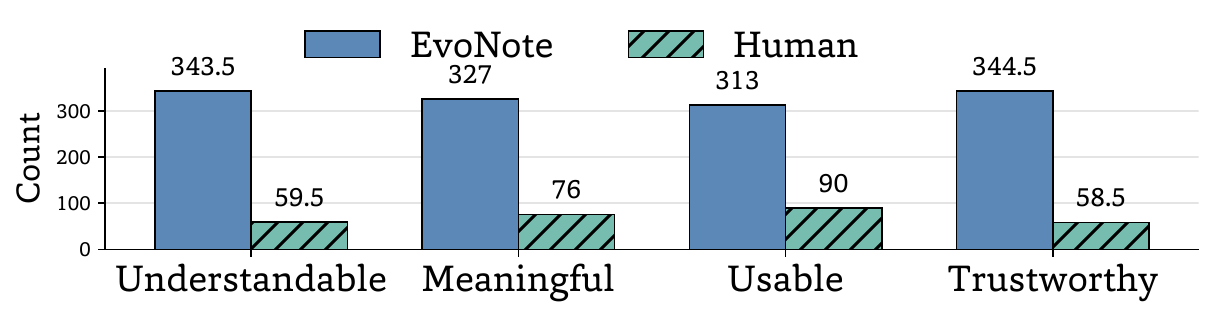}
\caption{
\textbf{Pairwise utility comparison} across health-specific utility dimensions on 403 cases where both human-written notes and \ours{} are judged \textit{Helpful}. 
Ties count as 0.5 for both sides.
}
    \label{fig:dimension_comparison}
\end{center}
\end{figure}

For judge-rated \textit{Helpful} human notes ($N=460$), \ours{} remains helpful in \textbf{87.6\%} of cases and is preferred in \textbf{74.1\%}.
Figure~\ref{fig:dimension_comparison} shows gains across all four health-specific utility dimensions, indicating broader improvements in public-facing correction quality.

\begin{table}[t]
\centering
\small
\resizebox{\columnwidth}{!}{%
\begin{tabular}{lcccc}
\toprule
\textbf{Prior correction logic} & \textbf{\#} & \textbf{\ours{}\textsc{-nm}} & \textbf{\ours{}} & \textbf{$\Delta$} \\
\midrule
No qualifying prior   & 38  & 86.84 & 86.84 & 0.00 \\
Prior $\leq$180 days  & 616 & 84.74 & 89.85 & +5.11 \\
Prior $>$180 days     & 546 & 85.16 & 89.47 & +4.30 \\
\bottomrule
\end{tabular}
}
\caption{\textbf{Pairwise win rate (\%) by prior correction-logic availability.}
We stratify \ds{} by whether the most similar qualifying correction logic occurs within 180 days, more than 180 days earlier, or is unavailable under the Q0.80 threshold defined in Appendix~\ref{app:logic-recurrence}. $\Delta$ denotes the gain of \ours{} over \ours{}\textsc{-nm}.}
\label{tab:prior-logic-gains}
\end{table}

\paragraph{Fine-grained evolving memory outperforms generic memory use.}
Among MedGemma-27B-based methods, all memory-based agents outperform the one-pass CrowdNotes+ baseline, confirming the value of reusable experience.

\begin{figure}[t]
\begin{center}
    \includegraphics[width=\linewidth]  {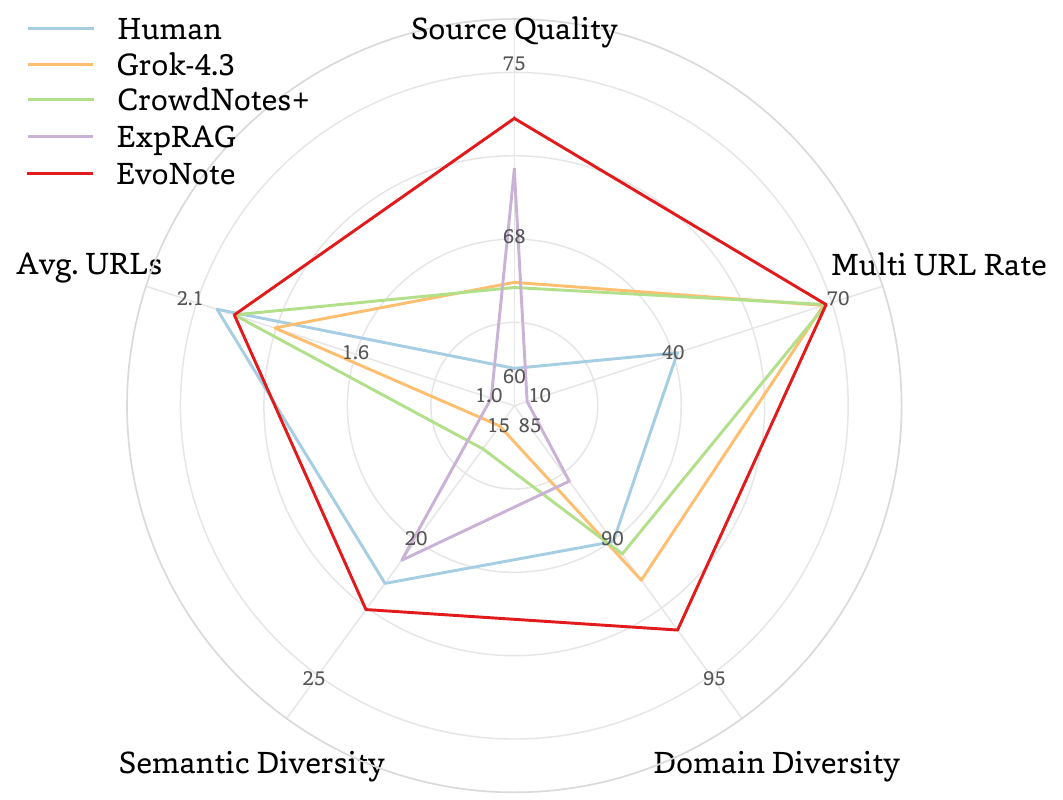}
\caption{\textbf{Evidence quality and diversity of \ours{} compared with representative baselines.} Metrics are defined in Appendix~\ref{app:evidence}.}
    \label{fig:quality-diversity}
\end{center}
\end{figure}

However, \ours{} achieves the strongest performance among memory-based methods.
Unlike ExpRAG, which retrieves full past trajectories, and ReMem, which asks the actor to infer reusable lessons from past trajectories, \ours{} explicitly distills judge feedback into phase-specific memory items for claim analysis, evidence acquisition, and writing.
This highlights the value of phase-directed memory (\S\ref{sec:memory-evolver}): reusable experience becomes most useful when tied to the correction stage that it should guide.

\paragraph{Both evolving memory and claim analysis matter.}
We conduct a series of ablations in order to isolate self-evolving memory and explicit claim analysis (see details in Appendix~\ref{app:variant-setup}):
\textbf{\ours{}\textsc{-fm}} freezes the warm-up memory bank during evaluation, \textbf{\ours{}\textsc{-nm}} removes memory retrieval and update, and \textbf{\ours{}\textsc{-nm-na}} further removes the Claim Analyzer.
Table~\ref{tab:main-results} shows that each removal degrades the performance, which confirms that \ours{} benefits from both accumulated experience and instance-specific verification planning.

\paragraph{Memory gains align with prior correction-logic availability.}
We further examine when evolving memory contributes to performance by stratifying \ds{} according to the availability of chronologically earlier posts with similar correction logic. Specifically, we use the Q0.80 empirical reference threshold that is defined in Appendix~\ref{app:logic-recurrence}. As shown in Table~\ref{tab:prior-logic-gains}, \ours{} and \ours{}\textsc{-nm} achieve the same pairwise win rate on the 38 cases with no qualifying prior neighbor.

When similar prior correction logic is available, \ours{} improves over \ours{}\textsc{-nm} by 5.11\% for neighbors within 180 days and by 4.30\% for neighbors beyond 180 days. These gains align evolving-memory benefits with reusable prior correction experience and remain substantial even when it occurred more than six months earlier.

\section{Discussion}
\label{sec:discussion}
\paragraph{\ours{} uses stronger and more diverse evidence.}
\label{sec:evidence-analysis}
We compare \ours{} with representative baselines using five evidence-side metrics: average cited URLs, multi-URL rate, source quality, cited-domain diversity, and semantic diversity (detailed in Appendix~\ref{app:evidence}). 
These metrics measure cross-source support, source credibility, and evidence diversity. 
Figure~\ref{fig:quality-diversity} shows that \ours{} achieves the strongest overall evidence profile, with more multi-source support, higher source quality, and stronger domain and semantic diversity. 
This suggests that self-evolving memory guides more deliberate evidence acquisition for later episodes.

\begin{table}[t]
\centering
\small
\resizebox{\columnwidth}{!}{%
\begin{tabular}{lcccccc}
\toprule
\textbf{Model} 
& \textbf{$\overline{T}$} 
& \textbf{Search} 
& \textbf{Visit} 
& \textbf{Infer.} 
& \textbf{Total} & \textbf{Win.} \\
\midrule
CrowdNotes+
& 6.0 & 4.8 & 23.5 & 17.9 & \ourcell{46.2} & \ourcell{69.21} \\
DeepResearch
& 13.7 & 12.8 & 71.3 & 68.9 & \ourcell{153.0} & \ourcell{78.09} \\
ExpRAG
& 8.6 & 6.1 & 34.7 & 30.2 & \ourcell{71.0} & \ourcell{79.63} \\
ReMem
& 15.6 & 5.4 & 47.6 & 65.3 & \ourcell{118.3} & \ourcell{74.50} \\
\midrule
\ours{}
& 12.4 & 7.9 & 55.1 & 49.6 & \ourcell{112.6} & \ourcell{\textbf{89.59}} \\
\bottomrule
\end{tabular}
}
\caption{\textbf{Time efficiency on \ds{} using a single H200 GPU.} We report mean turns $\overline{T}$, mean time (s) for search, visit, inference (note generation), mean total time, and Win rate over human-written notes.}
\label{tab:efficiency}
\vspace{-0.5em}
\end{table}

\paragraph{\ours{} improves quality with moderate cost.} 
As shown in Table~\ref{tab:efficiency}, \ours{} achieves the highest win rate while using fewer turns and less time than DeepResearch and ReMem. 
This suggests that phase-specific memory improves the quality-efficiency trade-off: \ours{} avoids the heavier information seeking of DeepResearch and the full-trajectory reasoning cost of ReMem by retrieving compact action-level strategies at the relevant stage.

\begin{table*}[t]
\centering
\small
\resizebox{\textwidth}{!}{%
\begin{tabular}{l
                cccc|
                cccc|
                cccc|
                cc}
\toprule
\multirow{2}{*}{\textbf{Model}} &
\multicolumn{4}{c|}{\textbf{\textsc{Text}}} &
\multicolumn{4}{c|}{\textbf{\textsc{Image}}} &
\multicolumn{4}{c|}{\textbf{\textsc{Video}}} &
\multicolumn{2}{c}{\textbf{Overall}} \\   
\cmidrule(lr){2-5} \cmidrule(lr){6-9} \cmidrule(lr){10-13} \cmidrule(lr){14-15}
& \textbf{R} & \textbf{C} & \textbf{H} & \textbf{Win.}
& \textbf{R} & \textbf{C} & \textbf{H} & \textbf{Win.}
& \textbf{R} & \textbf{C} & \textbf{H} & \textbf{Win.}
& \textbf{H} & \textbf{Win.} \\   
\midrule
Human Notes
& 78.00 & 51.00 & 42.00 & \ourcell{---}
& 85.00 & 60.00 & 48.00 & \ourcell{---}
& 88.00 & 63.00 & 46.00 & \ourcell{---}
& 45.33 & \ourcell{---} \\
Grok-4.3
& 74.00 & 66.00 & 49.00 & \ourcell{66.00}
& 78.00 & 68.00 & 59.00 & \ourcell{66.00}
& 72.00 & 55.00 & 47.00 & \ourcell{63.50}
& 51.67 & \ourcell{65.17} \\
CrowdNotes+
& \underline{89.00} & \underline{83.00} & \underline{65.00} & \ourcell{70.00}
& \underline{95.00} & \underline{80.00} & \underline{69.00} & \ourcell{\underline{68.00}}
& \underline{94.00} & 78.00 & 64.00 & \ourcell{63.50}
& \underline{66.00} & \ourcell{67.17} \\
ReMem
& \underline{89.00} & 74.00 & 57.00 & \ourcell{\underline{72.00}}
& 94.00 & 73.00 & 61.00 & \ourcell{66.00}
& 96.00 & \underline{84.00} & \underline{73.00} & \ourcell{\underline{74.50}}
& 63.67 & \ourcell{\underline{70.83}} \\
\midrule
\ours{}
& \textbf{95.00} & \textbf{88.00} & \textbf{78.00} & \ourcell{\textbf{81.50}}
& \textbf{100.00} & \textbf{96.00} & \textbf{87.00} & \ourcell{\textbf{86.00}}
& \textbf{96.00} & \textbf{86.00} & \textbf{81.00} & \ourcell{\textbf{83.00}}
& \textbf{82.00} & \ourcell{\textbf{83.50}} \\
\bottomrule
\end{tabular}}
\caption{\textbf{Effectiveness (\%) of \ours{} on 300 ``Needs More Ratings'' (NMR) Samples, evenly distributed across \textsc{Text}, \textsc{Image}, and \textsc{Video}.} 
``Human Baseline'' is the original human-written Community Notes.
Metrics are relevance after length-validity filtering (\textbf{R}), evidence correctness (\textbf{C}), note helpfulness (\textbf{H}), and pairwise win rate against the Human Baseline (\textbf{Win.}; ties count as 0.5).
See \S\ref{sec:exp-setup} for setup, and Table~\ref{tab:nmr-results-len} for length compliance rates.
Best and second-best results are \textbf{bolded} and \underline{underlined}.
}

\label{tab:nmr-results}
\end{table*}

\begin{figure}[t]
\begin{center}
    \includegraphics[width=0.95\linewidth]  {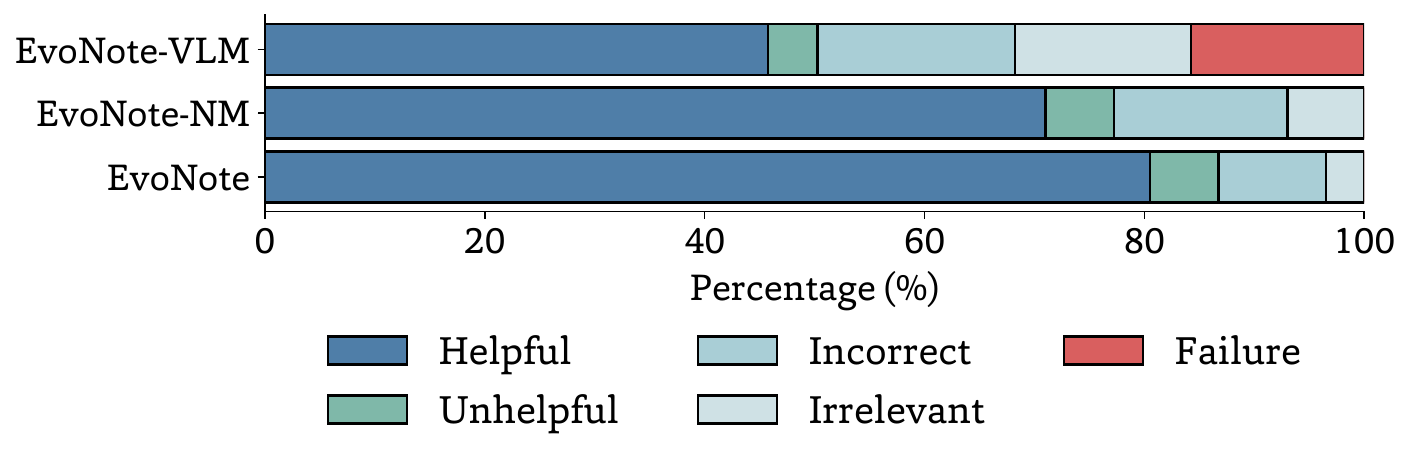}
    \caption{\textbf{VLM and caption-based LLM instantiations of \ours{}} on 400 \textsc{Image} cases.}
    \label{fig:vlm-limitation}
\end{center}
\end{figure}

\paragraph{\ours{} learns actionable health misinformation correction strategies.}
Across \ds{}, \ours{} accumulates 4,277 memory items, each tied to an action phase and reusable strategy.
Using a human-built attribution taxonomy and GPT-4.1 classification, we find that the dominant memories target primary-source verification, causal overgeneralization, safe next steps, and practical health implications (see \S\ref{app:memory-analysis}).

These phase-specific strategies address recurring correction failures.
Figure~\ref{fig:case-study} and \S\ref{app:case-study} further show one memory-guided case where retrieved experience improves evidence seeking and note framing over human notes, Grok-4.3, CrowdNotes+, and ReMem.

\paragraph{Captioning enables more reliable agentic health-related reasoning.} We examine whether \ours{} should directly instantiate its trajectory with a multimodal medical VLM. 
On the \textsc{Image} subset, \ours{} with a multimodal MedGemma-27B backbone fails to complete valid note-generation trajectories in nearly all cases; in fact, even after complex repair heuristics, 63 cases remain invalid (see Appendix~\ref{app:vlm-failure}). 
On this basis, Figure~\ref{fig:vlm-limitation} shows that the VLM version still underperforms the caption-based LLM pipeline, and it performs even below the no-memory LLM variant. 
This supports our captioning design: human-validated captions preserve correction-relevant visual content, while medical-capable LLMs offer more reliable tool use, evidence acquisition, and multi-step reasoning.

\paragraph{\ours{} remains effective before crowd consensus.}
We evaluate \ours{} on the most recent 300 \textit{Needs More Ratings} (NMR) cases, evenly split across \textsc{Text}, \textsc{Image}, and \textsc{Video}, where the helpfulness of existing human notes remains unresolved. 

The evaluation begins with the evolving memory accumulated from the 1,200 \ds{} instances, which is then updated over the NMR stream. All NMR cases occur chronologically after the \ds{} instances, ensuring that the initialized memory contains only prior correction experience and avoiding future-instance leakage. As shown in Table~\ref{tab:nmr-results}, \ours{} produces helpful notes for \textbf{82.0\%} of cases and achieves an \textbf{83.5\%} pairwise win rate over the existing human notes, with helpfulness rates of 78.0--87.0\% and win rates of 81.5--86.0\% across modalities. These results show that the benefits of \ours{} extend to posts awaiting crowd consensus, supporting timely correction while community ratings are still forming.

\section{Conclusion}
We presented \ours{}, a self-evolving agentic approach to health Community Notes generation. Our findings show that misinformation correction can become cumulative: recurring claims expose reusable correction experience, and agentic systems can improve by carrying that experience forward. On \ds{}, \ours{} outperforms human-written notes and strong automated baselines, with gains supported by stronger evidence use and actionable memory. This points to a broader design principle for misinformation governance, where each agentic correction trajectory should leave behind reusable experience for addressing the next similar claim. 

\section*{Acknowledgments}
This research is supported by the Ministry of Education, Singapore, under its Academic Research Fund Tier 1 (T1 251RES2508) and MOE AcRF TIER 3 Grant (MOE-MOET32022-0001), and by the NUS School of Computing Large Project Development Seed Grant (A-8002749-00-00).

\clearpage

\section*{Limitations}
Our presented work has shown that health misinformation correction can become cumulative by turning agentic correction trajectories into reusable memory. Several extensions of this framework remain open:

\paragraph{First, \ours{} focuses on English health misinformation.}
This setting is high-stakes and evidence-centric, making it suitable for studying reusable correction experience. Extending to political, financial, or sociocultural claims may require different utility criteria, as evidence can be more contested and context-dependent. Multilingual deployment also raises challenges in source coverage, local authority representation, and culturally grounded risk communication. Consequently, our findings might not generalize to regions with distinct linguistic and cultural profiles.
\paragraph{Second, \ours{} learns mainly from its own trajectories.}
Human-written notes are used for benchmarking and motivating analysis, while memory is distilled from agent-generated correction episodes. Future work could jointly learn from agent trajectories and high-quality human corrections. This may improve memory quality, but requires careful filtering because crowd-sourced signals can be noisy, incomplete, or affected by voting dynamics. Additionally, relying entirely on self-generated trajectories risks reinforcing or propagating the agent's initial systematic misconceptions across subsequent agentic learning episodes.
\paragraph{Third, self-evolution is implemented at the memory level.}
The backbone model, tools, and workflow remain fixed; adaptation comes from accumulating phase-specific action memories. This makes improvement interpretable and lightweight, but leaves broader forms of self-evolution, such as prompt revision, workflow adaptation, and tool-policy learning, for future work. As a result, the framework cannot correct inherent reasoning flaws present within the frozen core model.
\paragraph{Finally, \ours{} starts from already-flagged posts.}
We do not model upstream detection, prioritization, or real-time routing. Practical deployment should integrate self-evolving note generation with systems that identify emerging misinformation, estimate urgency, and decide when human review is needed.

\section*{Ethical Considerations}

\paragraph{Human Oversight in Health Correction.}
Health corrections can affect public understanding, risk perception, and health-related decisions. \ours{} should therefore be used as a human-augmenting system, not an autonomous arbiter of truth. Human reviewers should inspect cited evidence, verify medical claims, and decide whether a generated note is appropriate to surface. Without strict manual vetting, automated systems risk misinterpreting evolving scientific and clinical consensus.
\paragraph{Over-Trust and Automation Bias.}
Generated notes may appear authoritative even when evidence is incomplete or misrepresented. This risk is especially important in health contexts, where fluent explanations can hide subtle errors. Interfaces built around \ours{} should encourage active verification by exposing evidence snippets, highlighting uncertainty, and requiring source inspection before approval. Furthermore, clear visual markers should denote AI assistance to prevent users from defaulting to blind compliance.

\paragraph{Memory Safety.}
Self-evolving memory makes correction experience cumulative, which is a core strength of \ours{}. It can also propagate flawed lessons if incorrect trajectories are stored. Future deployments should include memory auditing, revision, decay, and deletion to prevent outdated or incorrect strategies from persisting. Malicious actors could also attempt to poison the system by intentionally spreading targeted claims to manipulate the accumulation of memory components.

\paragraph{Dual Use and Selective Evidence Use.}
Evidence-grounded note generation could be misused to produce persuasive but misleading corrections, cherry-pick sources, or suppress legitimate scientific disagreement. Mitigations should include transparent evidence presentation, source-diversity auditing, and safeguards against selective or adversarial use.

\paragraph{Data Use and Privacy.}
Our study uses public Community Notes data, public posts, and publicly accessible web evidence, without relying on private user information. Released benchmark resources should follow platform policies and be shared under research-oriented access conditions to support reproducibility while reducing privacy and misuse risks.

\bibliography{custom}
\clearpage

\appendix

\section{Details of Motivating Analysis}
\label{app:analysis}

This appendix provides details for the motivating analysis in \S\ref{sec:analysis}: including data filtering (\S\ref{app:analysis-data-filter}), claim pattern extraction (\S\ref{app:analysis-claim-pattern}), latency/visibility statistics (\S\ref{app:latency-visibility}), and cross-post correction-logic similarity analysis (\S\ref{app:logic-recurrence}).

\subsection{Health-Related Data Filtering}
\label{app:analysis-data-filter}

We collect publicly available, user-contributed Community Notes on X\footnote{\url{https://x.com/i/communitynotes/download-data}} up to 14 October 2025 and retain English entries for consistency.

In order to identify health-related notes, we follow the extraction pipeline of \citet{wu2026beyond}, which defines seven health-topic categories and validates a zero-shot filtering approach with Lingshu-32B \cite{xu2025lingshu}, a medical LLM. 
We adopt the same categories: \textbf{(1)} diseases or medical conditions, \textbf{(2)} drugs, vaccines, treatments, and tests, \textbf{(3)} public health guidance or policy, \textbf{(4)} wellness products, diets, and supplements, \textbf{(5)} healthcare professionals or systems, \textbf{(6)} biological or epidemiological concepts, and \textbf{(7)} health-related conspiracies or hoaxes.

We then use the following prompt against the Lingshu-32B LLM:

\begin{simple_prompt}{}
\small  

PROMPT = Classify whether the **core part** of the following text is health-related. Health-related means it substantively covers one or more of: \\
- Diseases or medical conditions \\
- Drugs, vaccines, treatments, procedures, tests \\
- Public health guidance or policy \\
- Wellness products, diets, supplements \\
- Healthcare professionals or systems \\
- Biological, virology, or epidemiology concepts \\
- Health-related conspiracies or hoaxes \\

Ignore casual or metaphorical mentions, non-health contexts, or isolated keywords. \\
Respond with yes or no on the last line. \\

Text: \{note\_text\}

\end{simple_prompt}

Next, we retrieve the associated flagged posts from X and remove unavailable posts or URL-only content. 
This yields 48,295 notes addressing 40,116 health-related flagged posts. 
Among them, 6,336 notes (13.12\%) have crowd-derived helpfulness labels, including 4,874 \textit{``Helpful''} notes and 1,462 \textit{``Not Helpful''} notes. 
The remaining cases are unresolved, i.e., \textit{``Needs More Ratings''}.

\subsection{Claim Pattern Extraction}
\label{app:analysis-claim-pattern}

We detail the correction-relevant claim taxonomy used in \S\ref{sec:analysis-setup}.
We derive the claim patterns from the 4,874 \textit{Helpful} human-written notes obtained in \S\ref{app:analysis-data-filter}, using them as high-precision signals for what made a flagged health post misleading as well as what evidence or framing made the correction useful.

Starting from recurring health misinformation patterns reported in prior work \cite{wang2019systematic,kington2021identifying,ngai2022impact,watson2024descriptions}, two human experts reviewed 200 sampled post-note pairs, summarized the main misleading mechanisms, and merged overlapping labels into the compact taxonomy in Table~\ref{tab:claim-patterns}.
Each category captures a recurring misleading mechanism, while its description notes the typical correction need revealed by helpful notes.

We then use GPT-4.1 \cite{openai2025gpt4_1} in order to classify the remaining helpful-note cases under this taxonomy.
For posts involving multiple mechanisms, we assign the primary pattern according to the correction strategy that is most central to the helpful note.
The resulting distribution, reported in Figure~\ref{fig:recur-analysis}, supports our analysis of recurring correction needs.

\begin{table}[t]
\centering
\small
\begin{tabular}{lccc}
\toprule
\textbf{Post Modality} & \textbf{Post-to-Note} & \textbf{Note-to-Verdict}  \\
\midrule
Text & 6.3 & 7.3  \\
Text \& Image & 7.1 & 6.6  \\
Text \& Video & 6.1 & 6.4  \\
\bottomrule
\end{tabular}
\caption{
\textbf{Median latency (in hours) for health Community Notes across post modalities.}
\textit{Post-to-Note} measures time from post creation to first note; \textit{Note-to-Verdict} measures time from note creation to crowd-derived helpfulness verdict.
}
\label{tab:note-delays}
\end{table}

\subsection{Latency and Visibility Statistics}
\label{app:latency-visibility}

We provide contextual statistics on the timeliness and coverage challenges of health Community Notes for correcting multimodal misleading posts, complementing the motivating analysis in \S\ref{sec:analysis}. We compute two latency metrics: \textbf{(1)} \textit{post-to-note latency}, the time from post creation to the first note, and \textbf{(2)} \textit{note-to-verdict latency}, the time from note creation to a crowd-derived helpfulness verdict, i.e., \textit{Helpful} or \textit{Not Helpful}. We also measure \textit{verdict coverage}, the percentage of notes that receive such a verdict.
We compute latency over the 6,336 health-related notes with crowd-derived helpfulness verdicts (obtained in \S\ref{app:analysis-data-filter}). 

As shown in Table~\ref{tab:note-delays}, median post-to-note latency ranges from 6.1 to 7.1 hours across modalities, while median note-to-verdict latency ranges from 6.4 to 7.3 hours. These statistics characterize two distinct sources of delay in the Community Notes pipeline: the time before a note is contributed and the subsequent time before a crowd-derived verdict is reached. The consistently multi-hour latencies across text-only, image-containing, and video-containing posts indicate that delayed correction is a challenge across modalities. During the first stage, a misleading post may circulate before any corrective note is available; during the second, a contributed note awaits sufficient community assessment to establish its helpfulness. Both stages therefore constrain timely health misinformation correction. This motivates accelerating evidence-grounded note generation, while also highlighting that faster generation alone does not resolve the downstream delay in crowd evaluation.

Note coverage is also limited. Among all 48,295 health-related notes, only 6,336 notes (13.12\%) receive a crowd-derived verdict, including 2,624 text-only cases, 2,207 text-image cases, and 1,505 text-video cases. The remaining 86.88\% stay unresolved as \textit{Needs More Ratings}. This large unresolved share highlights a gap between contributing a corrective note and obtaining the community assessment needed to determine its helpfulness. Moreover, only 4,874 notes, 10.1\% of all health-related notes, are ultimately rated \textit{Helpful}. These results show that many correction efforts remain delayed, unresolved, or never become visible as helpful corrections, echoing prior findings on Community Notes latency and coverage \cite{renault2024collaboratively,wu2026beyond}. This motivates LLM-augmented note generation approaches that can create evidence-grounded notes earlier while crowd consensus is still forming.

\begin{table}[t]
\centering
\small
\begin{tabular}{lrrrr}
\toprule
\textbf{Background} & \textbf{Sim.} & \textbf{Prior} & \textbf{$\leq$180d} & \textbf{$>$180d} \\
\textbf{percentile} & \textbf{threshold} & (\%) & (\%) & (\%) \\
\midrule
Q0.80 & 0.49 & 99.36 & 57.07 & 42.93 \\
Q0.85 & 0.52 & 98.89 & 57.08 & 42.92 \\
Q0.90 & 0.54 & 97.96 & 56.89 & 43.11 \\
Q0.95 & 0.59 & 94.47 & 56.89 & 43.11 \\
\bottomrule
\end{tabular}
\caption{\textbf{Prevalence and temporal distance of similar prior correction logic.} Similarity thresholds correspond to percentiles of the background distribution over 10,000 randomly sampled pairs of correction logics. The final two columns are computed among posts with a qualifying prior neighbor.}
\label{tab:correction-logic-recurrence}
\vspace{-1em}
\end{table}

\begin{table*}[h]
\centering
\small
\setlength{\tabcolsep}{10pt}
\begin{tabular}{@{}p{0.22\textwidth} p{0.16\textwidth} p{0.52\textwidth}@{}}
\toprule
\textbf{Pattern} & \textbf{Short Label\newline in Figure \ref{fig:recur-analysis}} & \textbf{Description} \\
\midrule
Misleading health statistics or quantitative evidence
& Misleading\newline Statistics
& The post uses numbers, rates, percentages, rankings, charts, counts, denominators, time windows, or quantitative comparisons in a misleading way. Corrections typically need to verify the original data source, denominator, baseline, time window, geographic scope, or whether the statistic supports the stated health claim. \\
\midrule

Exaggerated vaccine-risk\newline claims
& Vaccine Risk
& The post overstates, misattributes, or misrepresents harms, deaths, side effects, adverse events, or population-level risks from vaccines, especially COVID-19 vaccines. Corrections typically need to distinguish temporal association, adverse-event reporting, anecdotal cases, and causal evidence. \\
\midrule

Unsupported causal claims
& Unsupported\newline Causality
& The post claims or implies that one factor causes a health outcome, but the evidence does not support the causal link. Corrections typically need to check whether the claim improperly generalizes from anecdotes, correlations, isolated cases, temporal associations, or confounded observations. \\
\midrule

Hidden-cure or suppression narratives
& Suppression\newline Narrative
& The post claims that a cure, treatment, risk, agenda, or truth is being hidden, suppressed, covered up, or deliberately concealed by authorities, media, companies, experts, or elites. Corrections typically need to separate unsupported conspiracy framing from verifiable evidence about treatment efficacy, policy decisions, or public documentation. \\
\midrule

Decontextualized\newline institutional, policy, or\newline regulatory statements
& Institutional Context
& The post misrepresents the meaning, scope, timing, or implications of a statement, policy, ruling, approval, warning, or action by an institution, regulator, government, court, health authority, public official, or medical organization. Corrections typically need to restore the original institutional context and clarify what the statement or policy actually says. \\
\midrule

Misleading use of scientific evidence
& Scientific Evidence
& The post misuses, cherry-picks, overgeneralizes, or misreads scientific evidence, including studies, preprints, clinical data, expert consensus, official medical guidance, or evidence limitations. Corrections typically need to assess study design, evidence quality, clinical relevance, consensus, and whether the cited evidence actually supports the post's claim. \\
\midrule

Outdated or mismatched\newline media context
& Media Context
& The post uses an image, video, screenshot, quote, clip, transcript, headline, or event context in a misleading way, such as wrong time, place, person, source, original event, or visual implication. Corrections typically need to trace the media item back to its original source and surrounding context. \\
\midrule

Misrepresented medical\newline concept or biological\newline mechanism
& Medical Mechanism
& The post incorrectly explains a disease, diagnosis, symptom, biological process, ingredient, medical concept, mechanism, dosage, or technical health term. Corrections typically need to clarify the underlying medical or biological mechanism in accessible language. \\
\midrule

Misleading health advice or treatment efficacy claim
& Health Advice
& The post gives misleading advice about health behavior, prevention, self-care, diagnosis, nutrition, medication, supplement use, treatment effectiveness, or cure claims. Corrections typically need to verify whether the advice is evidence-based, safe, clinically appropriate, and not overstated beyond available evidence. \\
\midrule

Other
& Other
& The post does not fit any of the above recurring correction-relevant patterns. This label should be used sparingly, and the rationale should explain why the case is not primarily about statistics, causality, evidence use, institutional context, media provenance, medical mechanism, vaccine risk, suppression framing, or health advice. \\

\bottomrule
\end{tabular}
\caption{
\textbf{Correction-relevant health misinformation patterns.}
Two human annotators construct these categories from 200 sampled helpful-note cases; descriptions summarize the misleading mechanism and the corresponding misinformation correction need.
}
\label{tab:claim-patterns}
\end{table*}

\subsection{Cross-Post Correction Logic Similarity Analysis}
\label{app:logic-recurrence}

The analysis in \S\ref{sec:analysis-setup} characterizes recurrence using broad correction-relevant misinformation patterns. We complement this analysis at a finer granularity by examining whether later posts have chronologically earlier cases with similar verification and correction needs.

\paragraph{Correction logic extraction.}
We analyze the same 4,874 \textit{Helpful} notes used in \S\ref{sec:analysis-setup}. After deduplicating notes associated with the same flagged post, we obtain 3,779 unique posts. For each post and its multimodal content, GPT-4.1 extracts a concise \textit{correction logic} specifying (1) the health claim, (2) the reasoning error, missing context, or verification need, and (3) the evidence required to assess the claim. Three independent annotators evaluate 100 randomly sampled extractions, with 95\% judged correct by majority vote.

\paragraph{Prior correction-logic similarity.}
We encode each extracted correction logic with \texttt{sentence-transformers/all-mpnet-base-v2} and compute cosine similarity between embeddings. For each post, we identify the most similar correction logic among all chronologically earlier posts. To provide empirical reference points, we construct a background similarity distribution from 10,000 randomly sampled pairs of correction logics and consider its 80th, 85th, 90th, and 95th percentiles as reference thresholds. We label a prior neighbor as qualifying when its similarity exceeds the corresponding threshold. Table~\ref{tab:correction-logic-recurrence} reports the resulting coverage and temporal distance of these prior neighbors.

As each post is compared against a different number of chronologically earlier candidates, these thresholds serve as descriptive reference points rather than a formal statistical test of recurrence. Under these reference thresholds, 94.47--99.36\% of posts have a qualifying earlier neighbor. Approximately 43\% of the selected neighbors occur more than 180 days earlier. These results indicate that highly similar correction logic is frequently retrievable from the historical stream and can remain relevant across substantial temporal gaps.

We use this operational definition of prior correction-logic availability in \S\ref{sec:main-results} in order to examine how evolving-memory gains vary with the presence and temporal distance of relevant prior experience.

\clearpage

\section{Details of \ours{} Framework}
\label{app:framework}

This appendix details the main components of the \ours{} framework described in \S\ref{sec:method}. The detailed prompts are given in \S\ref{app:prompts}, and
a complete trajectory example is provided in \S\ref{app:demo}.

\subsection{Trajectory-Level Utility Judgment}
\label{app:social-utility-judge}

The \textbf{Social Utility Judge} agent implements the credit-assignment role introduced in \S\ref{sec:social-utility-judge}. 
Its role is to convert a completed note-generation trajectory into diagnostic feedback for future learning. 
Given the full trajectory, including claim analysis, search and visit actions, collected evidence, and the generated note, the Judge evaluates how the trajectory supports or weakens the final correction.

The Judge grounds this evaluation in four health-specific note qualities drawn from prior work in health communication. 
\textbf{Understandability} explains key distinctions or jargon in accessible language, helping readers grasp the correction without medical expertise \cite{sorensen2012health,warde2018plain,denniss2022development}. 
\textbf{Meaningfulness} clarifies why the correction matters for health interpretation, risk perception, or behavior, making its practical significance explicit \cite{ferrer2015risk,heydari2021effect,prike2023effective}. 
\textbf{Usability} provides safe and proportionate next-step guidance when user action may be affected, supporting responsible public decision-making \cite{denniss2022development,heydari2021effect}. 
\textbf{Trustworthiness} avoids overclaiming and acknowledges uncertainty or evidence limits where appropriate, preserving reliability under evolving medical evidence \cite{denniss2022development,schneider2021effects,kington2021identifying}.
These qualities define whether a correction is accessible to readers, useful for health interpretation, actionable when user behavior may be affected, and calibrated under evidence uncertainty.
The full Judge prompt is provided in Figure~\ref{prompt:social_utility_judge}.

The Judge produces two outputs. 
First, it provides fine-grained feedback that attributes note utility to trajectory decisions, such as claim analysis, evidence acquisition, uncertainty handling, or writing. 
Second, it assigns an overall utility mode, \textsc{Unsatisfactory}, \textsc{Satisfactory}, or \textsc{Excellent}, which determines how the trajectory should be converted into memory. 
This turns final-note evaluation into a structured learning signal for the Memory Evolver.

\subsection{Actionable Memory Evolution}
\label{app:memory-evolver}

The \textbf{Memory Evolver} agent implements the memory-construction role introduced in \S\ref{sec:memory-evolver}. 
Its role is to transform the Judge's feedback into reusable action-level memories. 
This step is necessary because raw trajectories contain many instance-specific details, such as URLs, query paths, and post-specific wording, that cannot be directly reused in future cases.

Given the trajectory and Judge outputs, the Evolver abstracts the episode into compact correction strategies. 
The full Memory Evolver prompt is provided in Figure~\ref{prompt:memory_evolver}. 
Each generated memory item follows the schema
\[
m = (\textsc{Trigger}, \textsc{ActionStrategy}, \textsc{Phase}).
\]
\textsc{Trigger} specifies when the memory should be retrieved, \textsc{ActionStrategy} states the reusable correction lesson, and \textsc{Phase} assigns the lesson to claim analysis, evidence acquisition, or note writing.

This structure makes memory directly actionable. 
Instead of storing full trajectories or generic reflections, \ours{} stores phase-specific guidance that can be retrieved at the point where it is useful. 
For example, a failed trajectory on a statistic-based health claim may yield an evidence-acquisition memory that triggers on similar numerical claims and recommends checking the original source, denominator, and surrounding context before writing.

\subsection{Memory-Guided Note Generation}
\label{app:memory-guided-generation}

Memory-guided generation implements the inference-time process introduced in \S\ref{sec:memory-guided-generation}. 
Its role is to use the current memory bank to guide a new note-generation episode without updating model parameters. 
The process is implemented through two prompt-driven agents: a \textbf{Claim Analyzer} agent and a \textbf{Note Writer} agent.

At each stage, \ours{} retrieves memories whose triggers match the current post or decision state and whose phase matches the current generation stage. 
The retrieved \textsc{ActionStrategy} fields guide the next step, while the final note remains grounded in evidence collected during the current episode. 
This allows prior experience to shape the trajectory while preserving instance-specific evidence verification.

\subsubsection{Claim Analyzer}
\label{app:claim-analyzer}

The \textbf{Claim Analyzer} performs the first memory-guided stage of note generation. 
Its role is to turn the flagged post into a focused verification brief before evidence acquisition begins. 
Given the post representation and retrieved claim-analysis memories, the Claim Analyzer identifies the central check-worthy claims, priority verification questions, and evidence needs.

The Claim Analyzer prompt is provided in Figure~\ref{prompt:claim_analyzer}. 
Its output follows a structured verification-brief schema, including the likely claim pattern, check-worthy claims, priority questions, evidence needs, correction focus, and relevant memory guidance. 
This brief defines what the Note Writer should verify and what kind of evidence it should seek.

The effect is to make evidence acquisition less reactive and more targeted. 
Instead of beginning from surface wording alone, the Note Writer starts from a memory-informed account of what must be checked, why it matters, and which recurring correction strategies may apply.

\subsubsection{Note Writer}
\label{app:note-writer}

The \textbf{Note Writer} executes the memory-guided correction trajectory. 
Its role is to acquire evidence and generate the final Community Note based on the verification brief, the current action history, collected evidence, and retrieved phase-matched memories.

At each step, the Note Writer selects one action from the available action space. 
It may issue a \textsc{Search} query, \textsc{Visit} a source returned by search or contained in the original post, \textsc{Write} a note when evidence is sufficient, or \textsc{Abstain} when the available evidence cannot support a reliable correction. 
The Note Writer prompt is provided in Figure~\ref{prompt:actor}. 
Retrieved memories guide query formulation, source prioritization, stopping decisions, uncertainty calibration, and note framing, but they are never treated as evidence themselves.

This design lets \ours{} reuse prior correction experience while maintaining evidence grounding for each new post. 
The resulting trajectory records the actions taken, the evidence collected, and the final note, which then becomes the next source of experience for the Social Utility Judge and Memory Evolver.

\section{\ds{} Benchmark Details}
\label{app:benchmark}

\subsection{Source Data Acquisition}
\label{app:benchmark-data}

We construct \ds{} by sampling 1,200 instances from the 6,336 health-related post--note pairs with crowd-derived verdicts obtained in \S\ref{app:analysis-data-filter}. 
In order to support controlled evaluation across modalities and note quality, we create three modality subsets (\textsc{Text}, \textsc{Image}, and \textsc{Video}), each containing 400 instances balanced between \textit{Helpful} and \textit{Not Helpful} human-written notes. 
This balanced design enables evaluation both on cases where existing human notes are useful and on cases where automated systems may provide better alternatives.

\begin{figure*}[t]
\begin{center}
    \includegraphics[width=\linewidth]{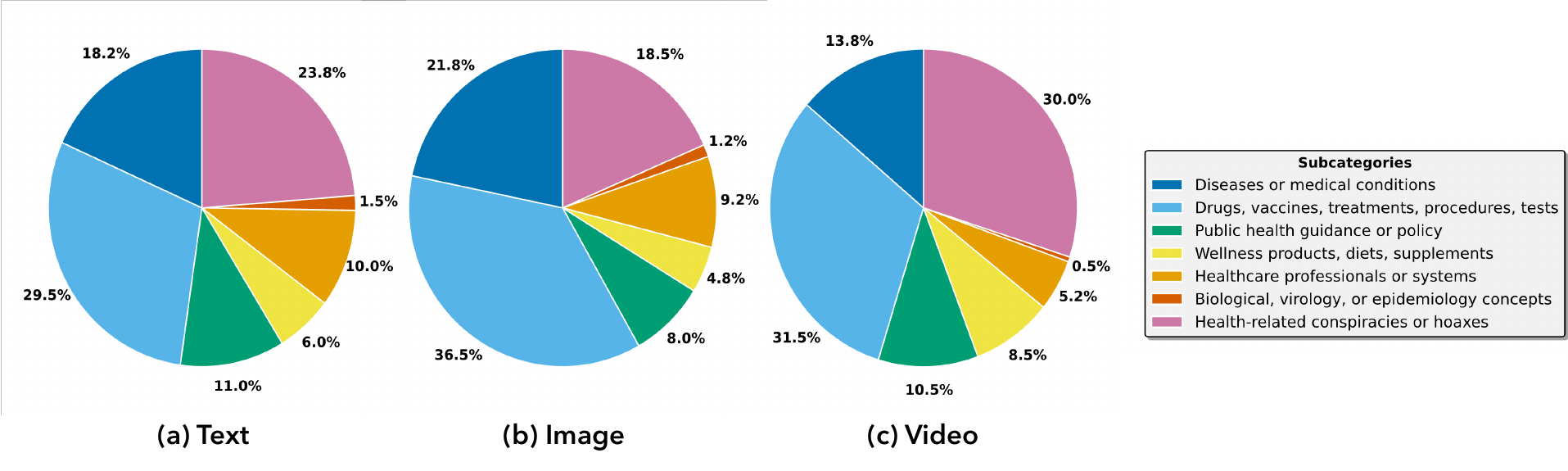}
    \caption{\textbf{Topic distribution of \ds{}.} Instances are grouped by post modality and assigned to representative health topics following prior work \cite{wu2026beyond}.}
    \label{fig:dataset-stats}
\end{center}
\end{figure*}

Figure~\ref{fig:dataset-stats} shows the topic distribution of \ds{}. 
Following prior work \cite{wu2026beyond}, we assign each instance to one health topic (\S\ref{app:analysis-data-filter}) using GPT-4.1 with temperature 0, based on the post text and, for image or video posts, the captioned multimodal content described in \S\ref{app:multimodal-captioning}. 
Across modalities, the benchmark covers diverse health misinformation topics, with drugs, vaccines, treatments, and medical conditions forming the largest shares, while video posts contain a higher proportion of conspiracy or hoax-related content. 
This distribution reflects the heterogeneous correction challenges faced by multimodal health Community Notes.

\begin{figure*}[t]
\begin{center}
    \includegraphics[width=\linewidth]  {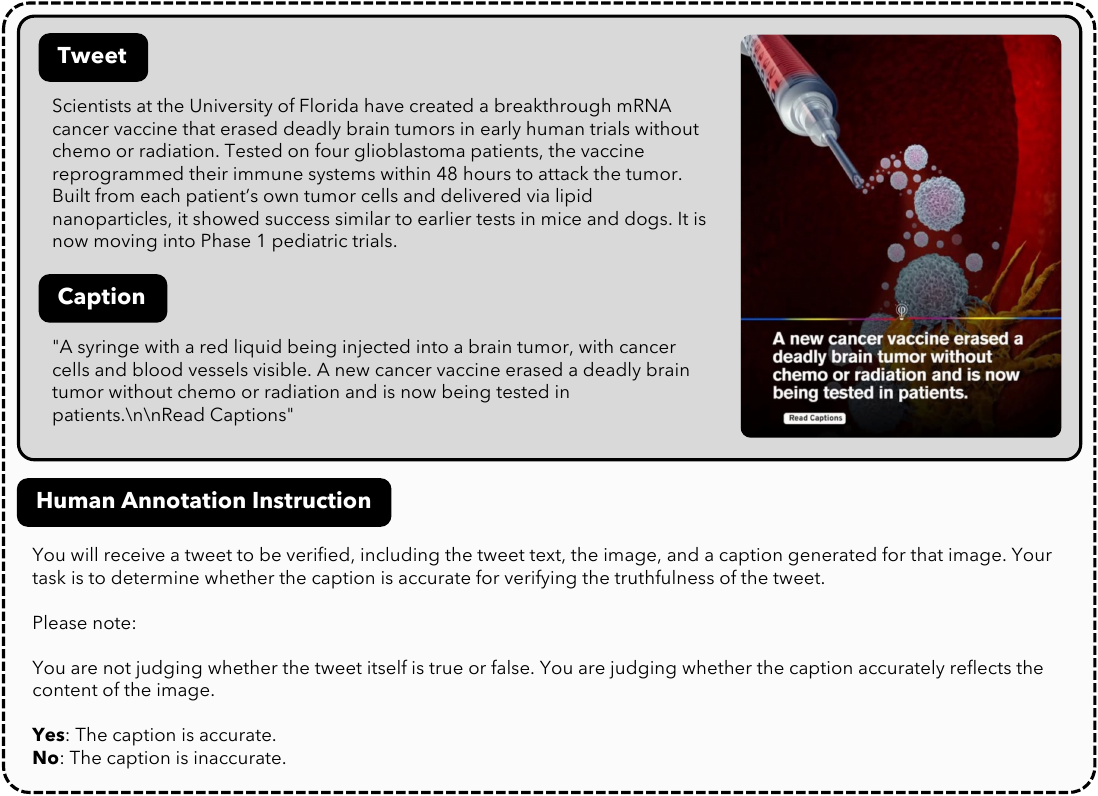}
    \caption{\textbf{Human validation protocol for multimodal caption quality} (\S\ref{app:multimodal-captioning}).}
    \label{fig:human-instruction-caption}
\end{center}
\end{figure*}

\subsection{Multimodal Post Preprocessing}
\label{app:multimodal-captioning}

We convert multimodal posts into textual representations so that image and video cases can share the same agentic note-generation pipeline as text-only cases. 
This design preserves the tool-use and multi-step reasoning strengths of LLM backbones while retaining correction-relevant visual information through captions. 
\S\ref{app:vlm-failure} further analyzes why direct multimodal-VLM instantiation is less reliable in this setting.

For image posts, we use Nanonets Docstrange\footnote{\url{https://github.com/NanoNets/docstrange}.}, which is designed for faithful extraction of visual and textual content from images. 
For video posts, we use Gemini-2.5-Flash~\cite{comanici2025gemini} in order to summarize correction-relevant visual content, on-screen text, and temporal context.
The resulting captions are concatenated with the original post text as the unified input representation for the model.

\subsection{Caption Quality Validation}
\label{app:caption-validation}

To verify that captioning preserves correction-relevant content, we conduct a human validation study on 200 randomly sampled multimodal instances from \ds{}, including 100 \textsc{Image} and 100 \textsc{Video} cases. 
As shown in Figure~\ref{fig:human-instruction-caption}, annotators view the original post and generated caption side by side, and judge whether the caption accurately reflects the multimodal content needed for verifying the post.

Three annotators independently label each instance. 
\textbf{For image captions, majority-vote accuracy is 0.95; for video captions, it is 0.96. 
Only 5 out of 100 cases show annotator disagreement in each setting.} 
These results indicate that caption-based representations reliably preserve the information needed for multimodal health misinformation correction.
\clearpage

\section{Hierarchical Note Utility Evaluation}
\label{app:eval-details}

This appendix details the hierarchical note utility evaluation scheme of \ds{} (\S\ref{sec:benchmark}).

Our evaluation proceeds in two stages. 
First, building on the human-validated helpfulness judgment protocol of \citet{wu2026beyond}, we determine \textbf{whether a note is \textit{Helpful}} through length compliance and progressive checks of evidence relevance, evidence correctness, and note-style helpfulness (\S\ref{app:helpfulness}). 
Second, for notes with the same binary helpfulness outcome, we compare their \textbf{public-facing social utility} through randomized pairwise judgment (\S\ref{app:pairwise-utility}).

\subsection{Hierarchical Helpfulness Judgment} 
\label{app:helpfulness}

A note is deemed \textit{Helpful} only if it passes the following four gates in a progressive order.

\textbf{Gate 1: Length Compliance.} 
    Community Notes must not exceed 280 characters, with each URL counted as one character\footnote{\url{https://docs.x.com/x-api/community-notes/quickstart}}; notes exceeding this are labeled \textit{Not Helpful} by a rule-based filter.
    
\textbf{Gate 2: Evidence Relevance.} 
    The cited evidence must provide factual context or clarification that helps readers evaluate the claim in the post.

\textbf{Gate 3: Evidence Representation Correctness.} 
    The note must accurately represent the cited sources without factual errors, exaggeration, or misleading framing, addressing a common risk in scientific and medical communication \cite{glockner2024missci,wuhrl2024understanding,glockner-etal-2025-grounding}.

\textbf{Gate 4: Note-Style Helpfulness.}
    Following Community Notes guidelines, the note must provide useful context that helps readers understand or critically evaluate the flagged post.

\paragraph{Gate Implementation.}
Gate 1 is implemented with the rule-based length filter above.
For Gates 2 and 3, we use the GPT-4.1-based prompts from the helpfulness judgment protocol of \citet{wu2026beyond}.
For Gate 4, we use \textsc{HealthJudge}, a Lingshu-7B-based binary helpfulness classifier fine-tuned on health Community Note post--note pairs by \citet{wu2026beyond}.
Since the Gates 2--4 judges are adopted without modification, we refer readers to \citet{wu2026beyond} for full prompt details.
Failure at any gate labels the note as \textit{Not Helpful}.
Judge reliability results are provided in \S\ref{app:helpfulness-reliability}.

\subsection{Pairwise Utility Comparison}
\label{app:pairwise-utility}

Binary helpfulness indicates whether a Community Note meets a minimum quality bar, but it does not distinguish between two Community Notes that are both helpful or that are both unhelpful.
We therefore further add pairwise utility comparison in order to capture which Community Note better serves public-facing factual health correction.
This complementary evaluation captures relative differences in how clearly and responsibly Community Notes communicate a correction.
For example, two helpful Community Notes may differ in how effectively they explain a misleading inference or how well they communicate evidence limitations.
Given two candidate Community Notes, a GPT-4.1-based judge compares them in a randomized order and first reasons over fine-grained health-communication criteria.
Randomizing the presentation order helps mitigate positional bias.
The judge then assigns a final outcome, \textit{win}, \textit{lose}, or \textit{tie}, based on the overall social utility of the two notes.

The criteria are grounded in established desiderata for health communication:
\emph{(1) understandability}, explaining key distinctions or necessary jargon in accessible language \cite{sorensen2012health,warde2018plain,denniss2022development},
\emph{(2) meaningfulness}, clarifying why the correction matters for health interpretation, risk perception, or behavior \cite{ferrer2015risk,heydari2021effect,prike2023effective},
\emph{(3) usability}, providing safe and proportionate next steps when the claim may affect user action \cite{denniss2022development,heydari2021effect},
and \emph{(4) trustworthiness}, avoiding overclaiming and acknowledging uncertainty or evidence limits where appropriate \cite{denniss2022development,schneider2021effects,kington2021identifying}.
Together, these dimensions capture both communicative effectiveness and responsible use of health-related evidence.
Their relevance depends on the claim: correcting a misrepresented study may primarily require clarification and qualification, whereas addressing an unsafe treatment claim may also require actionable guidance.

Pairwise preference remains a relative judgment: a winning note is better than its comparator under this protocol, but is not necessarily helpful in absolute terms.
Figure~\ref{prompt:pairwise_judgment} shows the prompt used for pairwise utility comparison.
We provide results on judge reliability in \S\ref{app:pairwise-reliability}.

\begin{figure*}
\begin{prompt}{Pairwise Judgment}
\small  

You are given a Tweet and two candidate Community Notes written by two different methods. \\

Tweet:
\{tweet text\}

Candidate Note 1:
\{Note 1\}

Candidate Note 2:
\{Note 2\} \\

Your task is to decide which note is better. \\

Evaluation criteria: \\
1. Understandable: explains key distinctions or necessary jargon in accessible language.\\
2. Meaningful: explains why the correction matters for health interpretation, risk perception, or behavior.\\
3. Usable: provides a safe and proportionate next step when the claim could affect user action. \\ 
4. Trustworthy: avoids overclaiming and acknowledges uncertainty or evidence limits where appropriate.
\\
\\
Instructions: \\
1. Compare the two notes criterion by criterion using the four criteria above. \\
2. For each criterion, choose exactly one of: Note 1, Note 2, or Tie. \\
3. A criterion should be marked Tie when neither note is clearly better on that criterion. \\
4. Then choose the better note overall, or choose Tie if neither note is clearly better overall. \\
\\
5. Use the following exact format for the criterion-level decisions, replacing CHOICE with exactly one of: Note 1, Note 2, or Tie. \\
Understandable: CHOICE \\
Meaningful: CHOICE \\
Usable: CHOICE \\
Trustworthy: CHOICE \\
\\
6. End your response with exactly one of the following lines: \\
Final decision: Note 1 \\
Final decision: Note 2 \\
Final decision: Tie \\

\end{prompt}
\caption{\textbf{Prompt for pairwise utility judgment} between two Community Notes candidates.}
\label{prompt:pairwise_judgment}
\end{figure*}
\clearpage

\subsection{Reliability of LLM Judges}
\label{app:llm-judge-reliability}

To assess the reliability of the LLM-based judgments used in evaluation, we validate the hierarchical helpfulness judges (\S\ref{app:helpfulness-reliability}) and the pairwise utility judge (\S\ref{app:pairwise-reliability}).

\subsubsection{Reliability of Hierarchical Helpfulness Judgment}
\label{app:helpfulness-reliability}

Length compliance is checked by a rule-based filter that directly follows the platform-enforced 280-character constraint.
For evidence relevance and evidence representation correctness, we use the GPT-4.1-based judges from \citet{wu2026beyond}, which were shown to achieve near-perfect human alignment on health Community Note evaluation.

We further test whether \textsc{HealthJudge}, the Gate 4 helpfulness classifier, generalizes to our multimodal setting.
We randomly sample 1,000 post-note pairs covering \textsc{Text}, \textsc{Image}, and \textsc{Video} cases, and compare \textsc{HealthJudge} with strong proprietary LLM judges, including GPT-4.1 and Claude-Sonnet-4.
As shown in Table~\ref{tab:helpfulness-reliability}, \textsc{HealthJudge} achieves the strongest performance, supporting its use as the final helpfulness gate after evidence relevance (Gate 2) and correctness (Gate 3) have been verified.

We refer readers to \citet{wu2026beyond} for the original human evaluation setup, prompt details, and \textsc{HealthJudge} training procedure.

\begin{table}[t]
\centering
\small
\begin{tabular}{lcc}
\toprule
\textbf{Model} & \textbf{Macro-F1 (\%)} & \textbf{Macro-Acc. (\%)} \\
\midrule
GPT-4.1 & 74.30 & 71.67 \\
Gemini-2.5-flash & 67.86 & 64.35 \\
Claude-Sonnet-4 & 74.58 & 72.40 \\
\midrule
\textsc{HealthJudge} & \textbf{89.21} & \textbf{88.46} \\
\bottomrule
\end{tabular}
\caption{\textbf{Effectiveness of \textsc{HealthJudge} for note helpfulness assessment}, validated by its superior performance on 1,000 unseen pairs of multimodal posts and corresponding notes (see \S\ref{app:helpfulness-reliability}).}
\label{tab:helpfulness-reliability}
\end{table}

\subsubsection{Reliability of Pairwise Utility Comparison}
\label{app:pairwise-reliability}

We validate the GPT-4.1-based pairwise utility judge (see \S\ref{app:pairwise-utility}) through two complementary human studies.

\begin{figure*}[t]
\begin{center}
    \includegraphics[width=\linewidth]{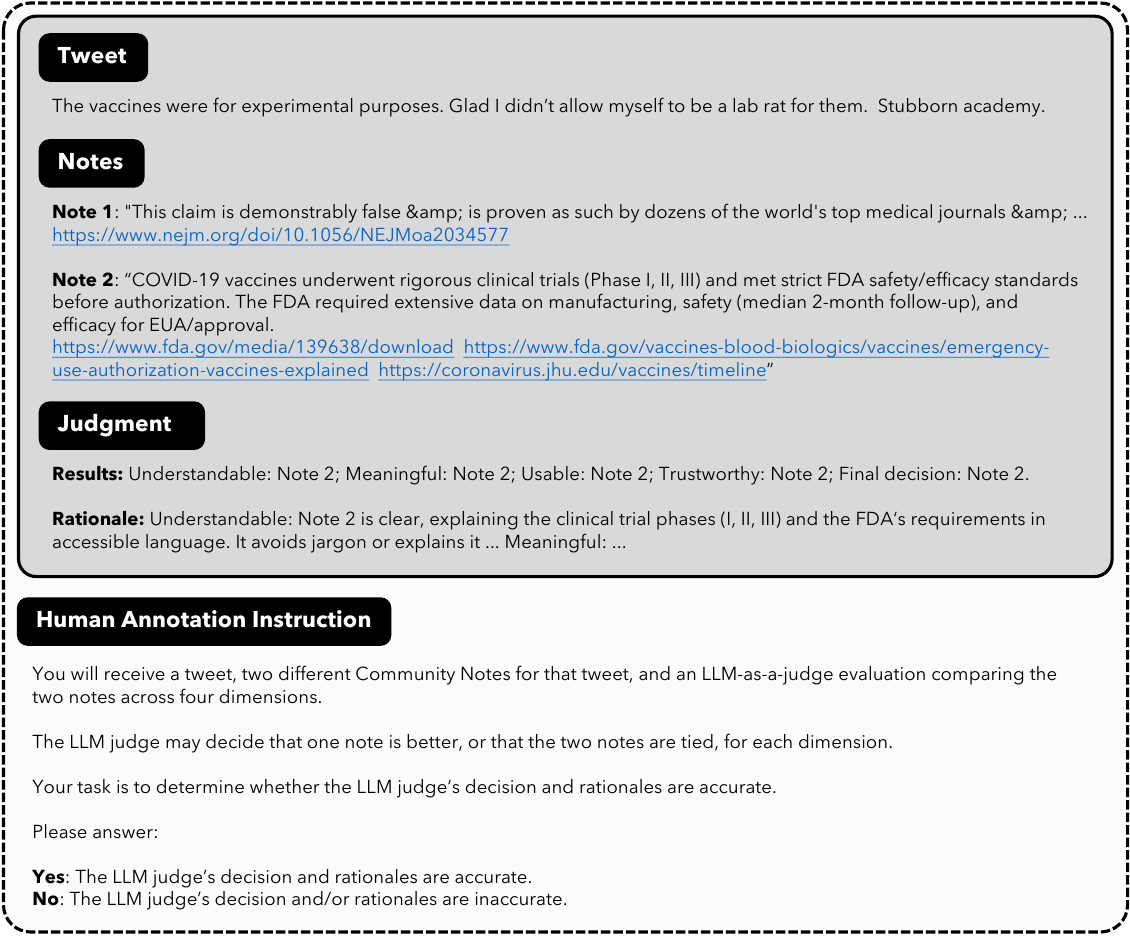}
    \caption{\textbf{Human audit protocol for GPT-4.1-based pairwise utility judgments.}
    Annotators review the candidate notes together with the judge's dimension-level decisions and rationales, and assess whether the comparison is accurate.}
    \label{fig:human-instruction-pairwise}
\end{center}
\end{figure*}

\paragraph{Rationale-aware human audit.}
We first conduct a human audit on 100 randomly sampled note pairs.
For each pair, three annotators review the flagged post, the two candidate notes, and the judge's dimension-level decisions and rationales, following the instruction in Figure~\ref{fig:human-instruction-pairwise}.

The annotators judge whether the comparison and its supporting rationales are accurate.
The human majority judges the pairwise comparison and its rationale as accurate in 98\% of cases, and only 4 of the 100 cases show inter-annotator disagreement.
This analysis assesses whether the judge's stated decisions and reasoning are substantively supported by the underlying notes.

\begin{table}[t]
\centering
\small
\resizebox{\columnwidth}{!}{%
\begin{tabular}{lcc}
\toprule
& \multicolumn{2}{c}{\textbf{Pairwise Utility Judge}} \\
\cmidrule(lr){2-3}
\textbf{Human majority} & \textsc{EvoNote} better & Human note better \\
\midrule
\textsc{EvoNote} better & 86 & 9 \\
Human note better       & 0  & 5 \\
\bottomrule
\end{tabular}
}
\caption{\textbf{Blind human validation of the GPT-4.1 based pairwise utility judge (detailed in \S\ref{app:pairwise-reliability}).}
The table compares the judge's overall preferences with independent human-majority judgments on 100 randomly sampled \textsc{EvoNote} and human-note comparisons.}
\label{tab:pairwise-blind-validation}
\end{table}

\paragraph{Independent blind human validation.}
We further conduct a blind evaluation to assess agreement without exposing annotators to the judge's outputs.
We randomly sample a separate set of 100 comparisons and recruit three annotators.
For each comparison, the two candidate notes are presented in randomized order and referred to only as Note 1 and Note 2.
System identities, LLM-as-a-Judge decisions, and rationales are hidden.
Each annotator independently selects Note 1, Note 2, or Tie for understandability, meaningfulness, usability, and trustworthiness, followed by an overall preference.
Both the LLM judge and the annotators may select Tie, although no tie appears as either the LLM judge's overall decision or the human-majority overall label in this sample.

Using the independent human-majority judgments as the reference, the LLM judge agrees on 91 of the 100 overall comparisons.
Table~\ref{tab:pairwise-blind-validation} gives the agreement matrix.
All nine disagreements occur in the same direction: the human majority prefers \textsc{EvoNote} in nine cases where the judge prefers the human-written note.
Notably, every judge's preference for \textsc{EvoNote} is also preferred by the human majority.
The LLM judge prefers \textsc{EvoNote} in 86\% of the sampled comparisons, compared with 95\% under the human-majority judgments.
This provides no indication of an \textsc{EvoNote}-favoring bias in this sample.

Moreover, dimension-level agreement is 93\% for understandability, 90\% for meaningfulness, 92\% for usability, and 90\% for trustworthiness.

\section{Experimental Setup}
\label{app:exp-setup}

This appendix supplements the experimental setup in \S\ref{sec:exp-setup} with details on baseline implementation (\S\ref{app:baselines}), shared implementation settings (\S\ref{app:implementation-details}), necessary evaluation constraints for fair comparison (\S\ref{app:eval-constraints}), and \ours{} variants used for controlled comparisons (\S\ref{app:variant-setup}).

\subsection{Baselines}
\label{app:baselines}

We compare \ours{} with three groups of baselines that cover progressively stronger forms of automated note generation.

\paragraph{Web-search-enabled LLMs.}
These baselines test whether strong general-purpose LLMs with external search can generate helpful notes without task-specific Community Notes design or memory:
\begin{itemize}[leftmargin=*]
    \item \textbf{GPT-4.1} \cite{openai2025gpt4_1},
    \item \textbf{Grok-4.3} \cite{xai2026grok43},
    \item \textbf{Gemini-2.5-Flash} \cite{comanici2025gemini}.
\end{itemize}

\paragraph{LLM-based Community Notes generation systems.}
These baselines represent task-adapted note-generation pipelines that explicitly target Community Notes-style correction:
\begin{itemize}[leftmargin=*]
    \item \textbf{CrowdNotes+} \cite{wu2026beyond}, based on MedGemma-27B \cite{sellergren2025medgemma} and instantiated under its Note Automation mode for automated evidence exploration; the evidence budget (top-$k$ evidence retrieved) is aligned with the number of evidences retrieved by \ours{} for a fair comparison.
    \item \textbf{DeepResearch} \cite{team2025tongyi}, a 30B open-source agent adapted as a search-intensive note-generation agent.
\end{itemize}

\paragraph{Memory-augmented LLM agents.}
These baselines test whether generic agent memory mechanisms can support reusable correction experience without the phase-specific memory design of \ours{}:
\begin{itemize}[leftmargin=*]
    \item \textbf{ExpRAG}, which retrieves prior trajectories as task-level context;
    \item \textbf{ReMem}, which integrates reasoning, action, and memory refinement for continual improvement.
\end{itemize}
Both are adapted from the Evo-Memory suite \cite{wei2025evo}.

\subsection{Implementation Details}
\label{app:implementation-details}

\paragraph{LLM Setup.}
For web-search-enabled LLM baselines, we evaluate GPT-4.1, Grok-4.3, and Gemini-2.5-Flash through the OpenRouter API\footnote{\url{https://openrouter.ai/}}, using each model's official web-search capability when available. We set \texttt{temperature}=0.0, \texttt{top\_p}=1.0, and \texttt{max\_tokens}=1024, and disable reasoning mode for Grok-4.3 and Gemini-2.5-Flash.

For LLM-based Community Notes generation systems, CrowdNotes+ follows the original implementation. DeepResearch uses \texttt{Alibaba-NLP/Tongyi-DeepResearch-30B-A3B} following its official technical report, with \texttt{temperature}=0.85, repetition penalty 1.1, \texttt{top\_p}=0.95, up to 128 tool calls, and the unchanged official prompt.

For memory-augmented agents, ExpRAG, ReMem, and \ours{} use \texttt{google/medgemma-27b-text-it} as the backbone LLM, deployed with vLLM on a single H200 GPU, with \texttt{temperature}=0.0 for all calls for stability. 

All experiments are run once.

\paragraph{Agent Configuration.}
For ExpRAG, ReMem, and \ours{}, we use the same execution budget: at most 15 steps, 5 searches, and 12 page visits.
Each episode requires at least one search and one visit before finalization, allows up to 5 write-refinement attempts, and is rerun up to 5 times upon execution failure.
Memory retrieval selects top-$k$ entries by lexical overlap with $k=4$.
We process the dataset with batch size 4 for memory updates.
For ReMem, we set the maximum memory-refinement loop to 4.

\paragraph{Search and Visit API Use.}
We use Serper\footnote{\url{https://serper.dev/}} for search and Firecrawl\footnote{\url{https://www.firecrawl.dev/}} for page visiting. Search returns up to 10 time-constrained results, deduplicates canonical URLs, and falls back to answer-box links when available. Visit extracts main-page markdown, retries failed or CAPTCHA-like pages with backoff and stealth proxy, and optionally falls back to Jina Reader\footnote{\url{https://jina.ai/}}. Failed calls return structured errors.

\paragraph{Community Notes Generation Prompts for Web-Search-Enabled LLMs.}
The prompt for web-search-enabled LLM baselines is shown below.

\begin{simple_prompt}{}
\small  

\textbf{TASK} \\
Write a factual, neutral Community Note for the following tweet, which may include attached images or videos. \\
 \\
\textbf{Tweet text}  \\
\{tweet\}\{multimodal\_caption\}  \\
 \\
\textbf{Hard constraints}  \\
- The note MUST be in English. \\
- The note MUST be exactly one single line. \\
- The note text MUST be concise and factual ($\le$ 270 characters, excluding citations). \\
- You MUST perform web search and cite reliable external sources using the default citation format. \\
- All referenced content MUST NOT be published later than **[{timestamp}]**. \\
- Be specific, objective, and verifiable. \\

\textbf{OUTPUT}  \\
Output only the note content, using the default citation format. Do not add explanations or extra text. \\

\end{simple_prompt}

\subsection{Evaluation Constraints}
\label{app:eval-constraints}
To ensure a fair and deployment-realistic evaluation, we impose the following constraints on all applicable methods:
\begin{itemize}[leftmargin=*]
    \item \textbf{Temporal search restriction.} 
    Web search results are constrained to content available up to the timestamp of the human-written note, preventing access to future information.

    \item \textbf{Chronological memory access.}
    For memory-based methods, episodes are processed in timestamp order, so each method can retrieve only experience accumulated from earlier episodes, i.e., note-generation trajectories for posts published earlier. This matches a practical streaming setting.
\end{itemize}

\subsection{\ours{} Variant Setup}
\label{app:variant-setup}
We design the following three key variants of \ours{} to assess the utility of memory and claim analysis:
\begin{itemize}[leftmargin=*]
\item \textbf{\ours{}\textsc{-fm}} (Frozen Memory): Allows retrieval from a fixed memory bank initialized with a held-out set that chronologically precedes \ds{} and is equivalent in size to 5\% of \ds{}. No new planner or actor memory is distilled or stored during evaluation. This ablation tests whether continual memory growth provides gains beyond retrieval from a fixed memory bank.
    \item \textbf{\ours{}\textsc{-nm}} (No Memory): disables both memory retrieval and memory update, while keeping the rest of the pipeline unchanged, including instance-specific rubric planning, budgeted evidence seeking, and refinement. 
    This tests whether memory contributes beyond the non-memory components of \ours{}.

    \item \textbf{\ours{}\textsc{-nm-na}} (No Memory, No Analyzer): disables memory and additionally removes the instance-specific Claim Analyzer. The actor directly performs evidence seeking and note writing from the post representation. 
    This tests whether explicit claim decomposition and verification planning improve note generation beyond a memory-free agentic pipeline.
\end{itemize}

\section{Additional Results and Analysis}
\label{app:experiments}

This appendix expands upon the quantitative and qualitative results (\S\ref{sec:main-results} and \S\ref{sec:analysis}), with analysis on note length compliance (\S\ref{app:length-compliance}), evidence use analysis (\S\ref{app:evidence}), memory composition and utility analysis (\S\ref{app:memory-analysis} and \S\ref{app:case-study}), and the validation of LLM-instantiated \ours{} for reliable task conduction (\S\ref{app:vlm-failure}).

\subsection{Note Length Compliance}
\label{app:length-compliance}

As described in \S\ref{app:helpfulness}, generated notes must satisfy the platform's 280-character limit before entering downstream helpfulness evaluation. 
Tables~\ref{tab:main-results-len} and~\ref{tab:nmr-results-len} report length compliance on the main benchmark and the \textit{Needs More Ratings} subset.

Length control is a major practical challenge for general web-search-enabled LLMs: GPT-4.1 and Gemini-2.5-Flash frequently exceed the platform limit, and Grok-4.3 also violates it on the \textit{Needs More Ratings} subset. 

In contrast, \ours{} achieves 100\% compliance across all modalities and evaluation settings, matching human notes and the strongest note-generation baselines. 
This shows that \ours{} maintains platform-valid note generation while improving helpfulness and utility.

\begin{table}[t]
\centering
\small
\resizebox{\columnwidth}{!}{%
\begin{tabular}{lccc|c}
\toprule
\textbf{Model} &
\textbf{\textsc{Text}} &
\textbf{\textsc{Image}} &
\textbf{\textsc{Video}} &
\textbf{Avg.} \\   
\midrule
Human Notes
& 100.00 & 100.00 & 100.00 & 100.00 \\
\midrule
GPT-4.1
& \textcolor{red}{46.25} & \textcolor{red}{41.00} & \textcolor{red}{36.00} & \textcolor{red}{41.08} \\
Grok-4.3
& \textcolor{red}{99.50} & \textcolor{red}{99.75} & \textcolor{red}{97.00} & \textcolor{red}{98.75} \\
Gemini-2.5-Flash
& \textcolor{red}{68.50} & \textcolor{red}{68.00} & \textcolor{red}{59.75} & \textcolor{red}{65.42} \\
\midrule
CrowdNotes+
& 100.00 & 100.00 & 100.00 & 100.00 \\
DeepResearch
& \textcolor{red}{98.50} & \textcolor{red}{97.50} & \textcolor{red}{94.25} & \textcolor{red}{96.75} \\
ExpRAG
& \textcolor{red}{98.50} & \textcolor{red}{98.00} & \textcolor{red}{97.25} & \textcolor{red}{97.92} \\
ReMem
& 100.00 & 100.00 & 100.00 & 100.00 \\
\midrule
\ours{}\textsc{-fm}
& 100.00 & 100.00 & 100.00 & 100.00 \\
\ours{}\textsc{-nm}
& 100.00 & 100.00 & 100.00 & 100.00 \\
\ours{}\textsc{-nm-na}
& 100.00 & 100.00 & 100.00 & 100.00 \\
\midrule
\ours{}
& \textbf{100.00} & \textbf{100.00} & \textbf{100.00} & \textbf{100.00} \\

\bottomrule
\end{tabular}}
\caption{\textbf{Note length compliance ratio (\%) of \ours{} on \ds{}.} 
``Human Baseline'' refers to original human-written Community Notes. 
\textbf{Avg. = average across text, image, and video} 
(see evaluation setup in \S\ref{sec:exp-setup}, \ours{} variants in \S\ref{sec:main-results}). Violations are marked in red.}
\label{tab:main-results-len}
\end{table}
\begin{table}[t]
\centering
\small
\resizebox{\columnwidth}{!}{%
\begin{tabular}{lcccc}
\toprule
\textbf{Model} & \textbf{\textsc{Text}} & \textbf{\textsc{Image}} & \textbf{\textsc{Video}} & \textbf{Avg.} \\
\midrule
Human Notes
& 100.00 & 100.00 & 100.00 & 100.00 \\
Grok-4.3
& \textcolor{red}{76.00} & \textcolor{red}{85.00} & \textcolor{red}{72.00} & \textcolor{red}{77.67} \\
CrowdNotes+
& 100.00 & 100.00 & 100.00 & 100.00 \\
ReMem
& \textcolor{red}{98.00} & \textcolor{red}{99.00} & 100.00 & \textcolor{red}{99.00} \\
\midrule
\ours{}
& \textbf{100.00} & \textbf{100.00} & \textbf{100.00} & \textbf{100.00} \\
\bottomrule
\end{tabular}}
\caption{\textbf{Note length pass ratio (\%) on 300 ``Needs More Ratings'' (NMR) Samples, evenly distributed across \textsc{Text}, \textsc{Image}, and \textsc{Video}.} 
\textbf{Avg.} denotes the average across the three modalities. Violations are marked in red.}
\label{tab:nmr-results-len}
\end{table}

\subsection{Evidence Use Analysis}
\label{app:evidence}

This appendix details the evidence-use analysis in \S\ref{sec:evidence-analysis}. 
Since health Community Notes are only useful when their corrections are grounded in reliable and sufficiently contextualized evidence, we evaluate not only whether a note cites sources, but also \textit{what kinds of sources} it cites and whether the cited evidence is diverse enough to support cross-checking. 
We therefore measure evidence use along five dimensions: source quality, domain diversity, semantic diversity, average number of cited URLs, and multi-URL rate.

\paragraph{Source Quality.}
We measure the credibility of cited evidence at the domain level. 
For each note $i$, we extract all URLs, convert them to registrable domains using eTLD+1, and deduplicate them into a domain set $\mathcal{D}_i$. 
Each domain $d$ is mapped to a quality tier $\tau(d) \in \{1,2,3,4\}$ using our maintained domain-tier table, with unseen domains assigned to the lowest tier. 
\textbf{Tier 4} includes official, primary, or highly institutionalized sources, such as government agencies, public-health organizations, authoritative medical resources, and established fact-checking organizations. 
\textbf{Tier 3} includes peer-reviewed venues, medical databases, universities, research institutes, teaching hospitals, and professional societies. 
\textbf{Tier 2} includes mainstream news outlets, science media, reference sites, preprint servers, and general nonprofit sources. 
\textbf{Tier 1} includes unknown sources, forums, blogs, marketing sites, user-generated-content platforms, and other weakly verifiable sources. 
Representative examples are shown in Table~\ref{tab:tier_examples}.

\begin{table}[t]
\centering
\small
\begin{tabular}{cllp{0.43\columnwidth}}
\toprule
\textbf{Tier} & \textbf{Domain} & \textbf{Count} & \textbf{Source Type} \\
\midrule
4 & nih.gov & 1,716 & public health agency \\
4 & cdc.gov & 1,349 & government/regulator \\
4 & who.int & 574 & intl public health \\
\addlinespace
3 & bmj.com & 105 & journal/publisher \\
3 & heart.org & 47 & professional society \\
3 & aap.org & 46 & professional society \\
\addlinespace
2 & reuters.com & 380 & news media \\
2 & bbc.com & 292 & news media \\
2 & apnews.com & 229 & news media \\
\addlinespace
1 & x.com & 192 & social media platform \\
1 & youtube.com & 83 & UGC/social platform \\
1 & reddit.com & 60 & UGC/social platform \\
\bottomrule
\end{tabular}
\caption{Representative examples by source-quality tier.}
\label{tab:tier_examples}
\end{table}

We convert each tier into a log-scaled quality score:
\[
q(d) = 100 \cdot \frac{\log \tau(d)}{\log 4}.
\]
Thus, Tier 1--4 domains receive scores of $0$, $50$, $79.25$, and $100$, respectively. 
The note-level source quality score is:
\[
S_{\mathrm{src}}(i) =
\begin{cases}
\frac{1}{|\mathcal{D}_i|}\sum_{d \in \mathcal{D}_i} q(d), & |\mathcal{D}_i| > 0, \\
0, & |\mathcal{D}_i| = 0.
\end{cases}
\]
We report the dataset-level average:
\[
S_{\mathrm{src}} = \frac{1}{N}\sum_{i=1}^{N} S_{\mathrm{src}}(i).
\]

\paragraph{Domain Diversity.}
Source quality alone does not capture whether a note triangulates evidence across independent sources. 
We therefore measure how evenly a note distributes citations across domains. 
For each note $i$, let $\mathcal{U}_i$ be the multiset of cited URLs with parsable domains, $n_i = |\mathcal{U}_i|$, and $p_i(d)=c_i(d)/n_i$ the citation distribution over domains. 
For notes with $n_i \leq 1$, we set the score to 0. 
Otherwise, we compute normalized entropy:
\[
S_{\mathrm{dom}}(i) =
100 \cdot
\frac{-\sum_d p_i(d)\log p_i(d)}{\log n_i}.
\]
This score reaches 100 when all cited URLs come from distinct domains and decreases when citations concentrate on the same source. 
The dataset-level score is averaged over notes with at least two effective URLs:
\begin{equation*}
S_{\mathrm{dom}} =
\frac{1}{|\mathcal{I}_{\geq 2}|}
\sum_{i \in \mathcal{I}_{\geq 2}} S_{\mathrm{dom}}(i),
\end{equation*}
\begin{equation*}
\mathcal{I}_{\geq 2}=\{i:n_i\geq2\}
\end{equation*}

\paragraph{Semantic Diversity.}
We would like to note that domain diversity may still overestimate evidence breadth when different sources repeat the same information. 
We therefore measure semantic diversity among cited evidence pages. 
For each note $i$, we embed the corresponding extracted content of each cited URL using \texttt{jinaai/jina-embeddings-v3}. 
Let $\mathcal{E}_i=\{\mathbf{e}_{i1},\ldots,\mathbf{e}_{iM_i}\}$ denote valid evidence embeddings. 
If $M_i<2$, we set the score to 0. 
Otherwise, we compute the average pairwise cosine distance as follows:
\begin{equation*}
S_{\mathrm{sem}}(i) =
100 \times
\frac{2}{M_i(M_i-1)} \times
\end{equation*}
\begin{equation*}
\sum_{1 \leq a < b \leq M_i} 
\frac{1-\cos(\mathbf{e}_{ia},\mathbf{e}_{ib})}{2}
\end{equation*}

Higher scores indicate that cited evidence is less semantically redundant. 
The dataset-level score is averaged over notes with at least two valid evidence embeddings:

\begin{equation*}
S_{\mathrm{sem}} =
\frac{1}{|\mathcal{I}_{\geq 2}|}
\sum_{i \in \mathcal{I}_{\geq 2}} S_{\mathrm{sem}}(i),
\end{equation*}
\begin{equation*}
\mathcal{I}_{\geq 2}=\{i:M_i\geq2\}.
\end{equation*}

\paragraph{Average URLs and Multi-URL Rate.}
We also report two citation-coverage metrics. 
For each note $i$, let $u_i$ be the number of extracted URL citations. 
The average number of cited URLs is
\[
\mathrm{AvgURLs} = \frac{1}{N}\sum_{i=1}^{N} u_i .
\]
The multi-URL rate measures the fraction of notes that cite at least two URLs:
\[
\mathrm{MultiURLRate} = \frac{1}{N}\sum_{i=1}^{N}\mathbb{I}[u_i \geq 2].
\]
Overall, the above metrics quantify whether a method grounds notes in external evidence as well as how often it supports a correction with multiple sources.

\begin{table*}[t]
\centering
\small
\resizebox{\textwidth}{!}{%
\begin{tabular}{lccccc}
\toprule
\textbf{Model} 
& \textbf{Source Quality} 
& \textbf{Domain Diversity} 
& \textbf{Semantic Diversity} 
& \textbf{Avg. URLs} & \textbf{Multi. URL Rate}\\
\midrule
Human Notes
& 61.68 & 90.04 & 21.59 & \textbf{2.03} & 40.83 \\
Grok-4.3
& 65.55 & 91.47 & 15.75 & 1.83 & 68.42 \\
CrowdNotes+
& 65.32 & 90.50 & 16.60 & 1.97 & 68.92 \\
ExpRAG
& 70.63 & 87.80 & 20.73 & 1.08 & 12.42 \\
\midrule
\ours{}
& \textbf{72.93} & \textbf{93.32} & \textbf{22.56} & 1.97 & \textbf{68.92} \\
\bottomrule
\end{tabular}
}
\caption{
\textbf{Evidence quality and diversity of \ours{} vs. representative baselines on \ds{}.}
We compare cited evidence across source quality, domain diversity, semantic diversity, average cited URLs, and multi-URL rate. Metric formulations are detailed in \S\ref{app:evidence}.
}
\label{tab:quality-diversity}
\end{table*}

\paragraph{Results.}
Table~\ref{tab:quality-diversity} and Figure~\ref{fig:quality-diversity} show that \ours{} achieves the strongest overall evidence profile among representative baselines. 
It obtains the highest source quality score, domain diversity, and semantic diversity, while matching the strongest multi-URL rate and maintaining nearly the same average number of cited URLs as human notes and CrowdNotes+. 
This indicates that \ours{} does not improve evidence use by simply citing more URLs. 
Instead, it retrieves stronger sources and combines less redundant evidence, suggesting that evolving experience memory helps the agent learn when to seek primary sources, verify source-claim alignment, and broaden the evidence context.

\subsection{Memory Utility Analysis}
\label{app:memory-analysis}

We further analyze the evolving experience memory to understand what reusable strategies \ours{} learns from its own correction trajectories. 
This analysis complements \S\ref{sec:analysis} by examining the composition of Note Writer memories, i.e., memories used during evidence acquisition and note writing.

Across \ds{}, \ours{} accumulates 4,277 memory items, each associated with an action phase and a reusable strategy. 
To characterize these memories, two human annotators collaboratively reviewed 200 sampled memory items and constructed an attribution taxonomy. 
We then use GPT-4.1 to classify the remaining memory items by their primary attribution category. 
Table~\ref{tab:attribution_ratio_description} reports the resulting distribution and category descriptions.

The learned memories concentrate on concrete correction operations. 
For evidence acquisition, the largest categories are primary or official-source verification, causal overgeneralization checks, evidence-quality assessment, and domain-specific verification. 
For note writing, the dominant categories involve safe next-step guidance, practical health implications, contextualization, and uncertainty calibration. 
These observed patterns show that \ours{} does not merely store past cases; instead, it distills useful recurring correction failures into phase-specific strategies that can eventually guide later evidence seeking and note framing.
\begin{table*}[t]
\centering
\scriptsize
\setlength{\tabcolsep}{3pt}
\renewcommand{\arraystretch}{1.12}

\begin{tabularx}{\textwidth}{
>{\raggedright\arraybackslash}p{0.15\textwidth}
>{\raggedright\arraybackslash}X
r
r
>{\raggedright\arraybackslash}p{0.18\textwidth}
>{\raggedright\arraybackslash}X
}
\toprule
\textbf{Failure Category} &
\textbf{Example Pattern} &
\textbf{Count} &
\textbf{\%} &
\textbf{Repairability} &
\textbf{System Consequence} \\
\midrule

Ungrounded final answer
& \texttt{\{"action":"write", "support":[]\}}
& 36
& 57.1
& Not repairable; required evidence is absent
& Rejected due to missing evidence URLs \\

Length-control failure
& Valid \texttt{\{"text": ..., "reason": ...\}} but output remains too long
& 18
& 28.6
& Not repairable without regenerating content
& Cannot produce platform-compliant note \\

Residual schema mismatch
& Missing \texttt{reason}, missing \texttt{action}, or missing discriminator
& 6
& 9.5
& Partially repairable in principle, but unresolved here
& Rejected by schema validation \\

Runtime timeout
& \texttt{httpx.ReadTimeout}
& 3
& 4.8
& Not applicable
& Episode interrupted \\

\midrule
\textbf{Total failed runs}
& --
& \textbf{63}
& \textbf{100.0}
& --
& -- \\
\bottomrule
\end{tabularx}

\caption{
\textbf{Remaining agentic execution failures of the direct multimodal-LLM-instantiated \ours{} on 400 \textsc{Image} cases in \ds{} after automatic repair.}
Percentages are computed over the 63 failed runs. See \S\ref{app:vlm-failure} for more detailed analyses.
}
\label{tab:direct-vlm-failures}
\end{table*}

\begin{table*}[htbp]
\centering
\small
\begin{tabularx}{\textwidth}{l c X}
\toprule
\textbf{Attribution} & \textbf{Ratio} & \textbf{Description} \\
\midrule
\multicolumn{3}{l}{\textbf{Evidence Acquisition Stage}} \\
\midrule

PRIMARY\_OFFICIAL\_VERIFICATION
& 21.0\%
& Verify concrete factual, policy, legal, governmental, electoral, regulatory, or official-action claims using primary or authoritative sources. \\

CAUSAL\_GENERALIZATION
& 20.4\%
& Check whether the post improperly generalizes from anecdotes, single incidents, isolated cases, correlations, or temporal associations to causal or population-level conclusions. \\

EVIDENCE\_QUALITY\_CONSENSUS
& 16.4\%
& Assess scientific evidence quality, methodological validity, clinical evidence, or authoritative consensus, especially for health, medicine, public health, science, safety, or technical claims. \\

DOMAIN\_SPECIFICITY
& 14.5\%
& Verify domain-specific details such as medical procedures, treatments, ingredients, side effects, diagnoses, guidelines, product claims, technical definitions, or condition-specific details. \\

SOURCE\_ALIGNMENT
& 10.9\%
& Check whether the linked/cited source actually supports the post's claim or implied broader conclusion. Use this when the memory teaches the actor to inspect a linked article, cited study, headline, screenshot, or source and verify whether it really backs the claim. \\

QUANT\_DATA\_CONTEXT
& 9.5\%
& Verify statistics, rates, percentages, graphs, charts, comparisons, denominators, baselines, time windows, geographic scope, or other quantitative context. \\

MEDIA\_QUOTE\_PROVENANCE
& 5.3\%
& Trace videos, images, screenshots, quotes, clips, transcripts, or public-figure statements back to their original source and surrounding context. \\

CLAIM\_SCOPE\_AND\_CONTEXT
& 1.9\%
& Ensure the note explicitly targets the right claim scope, implied claim, category distinction, source limitation, or context. Use this when the memory is more about how to phrase the note than about what to search. \\

\midrule
\multicolumn{3}{l}{\textbf{Note Writing Stage}} \\
\midrule

SAFE\_NEXT\_STEP 
& 30.9\% 
& Add safe, general next-step guidance, such as consulting qualified professionals, checking trusted public-health sources, or avoiding self-diagnosis or unsafe action. \\

PRACTICAL\_IMPLICATION 
& 24.6\% 
& Add why the correction matters for risk interpretation, real-world understanding, reader decision-making, or practical consequences. \\

CONTEXT\_IMPACT 
& 22.8\% 
& Add background, consequences, real-world implications, institutional context, policy context, historical context, or broader framing that helps readers understand the claim. \\

UNCERTAINTY\_LIMITS 
& 11.3\% 
& Calibrate uncertainty, evidence strength, study limitations, methodological limits, data gaps, or avoid overclaiming beyond available evidence. \\

MECHANISM\_EXPLANATION 
& 8.5\% 
& Add a short explanation of the mechanism, reasoning, causal pathway, definition, or underlying issue behind the correction. \\

TONE\_SCOPE\_CLARITY 
& 1.5\% 
& Improve neutrality, concision, clarity, scope control, category distinctions, terminology, or avoid speculation, emotional wording, misleading phrasing, or overgeneralization. \\

STRENGTHEN\_OR\_UPDATE 
& 0.4\% 
& Find stronger sources, official updates, current status, primary context, media context, or additional data limitations to support a better note. Use this when the memory is mainly about additional searching or sourcing rather than wording. \\

\bottomrule
\end{tabularx}
\caption{
\textbf{Attribution categories of 4,277 Note Writer memories obtained by \ours{} from \ds{}.}
Each memory is assigned to its primary reusable correction strategy, separately for evidence acquisition and note writing. Detailed analysis is in \S\ref{app:memory-analysis}.
}
\label{tab:attribution_ratio_description}
\end{table*}

\subsection{Case Study: Retrieved Memory Guides Regulatory-Process Verification}
\label{app:case-study}
\begin{figure*}[t!]
\begin{center}
    \includegraphics[width=\linewidth]  {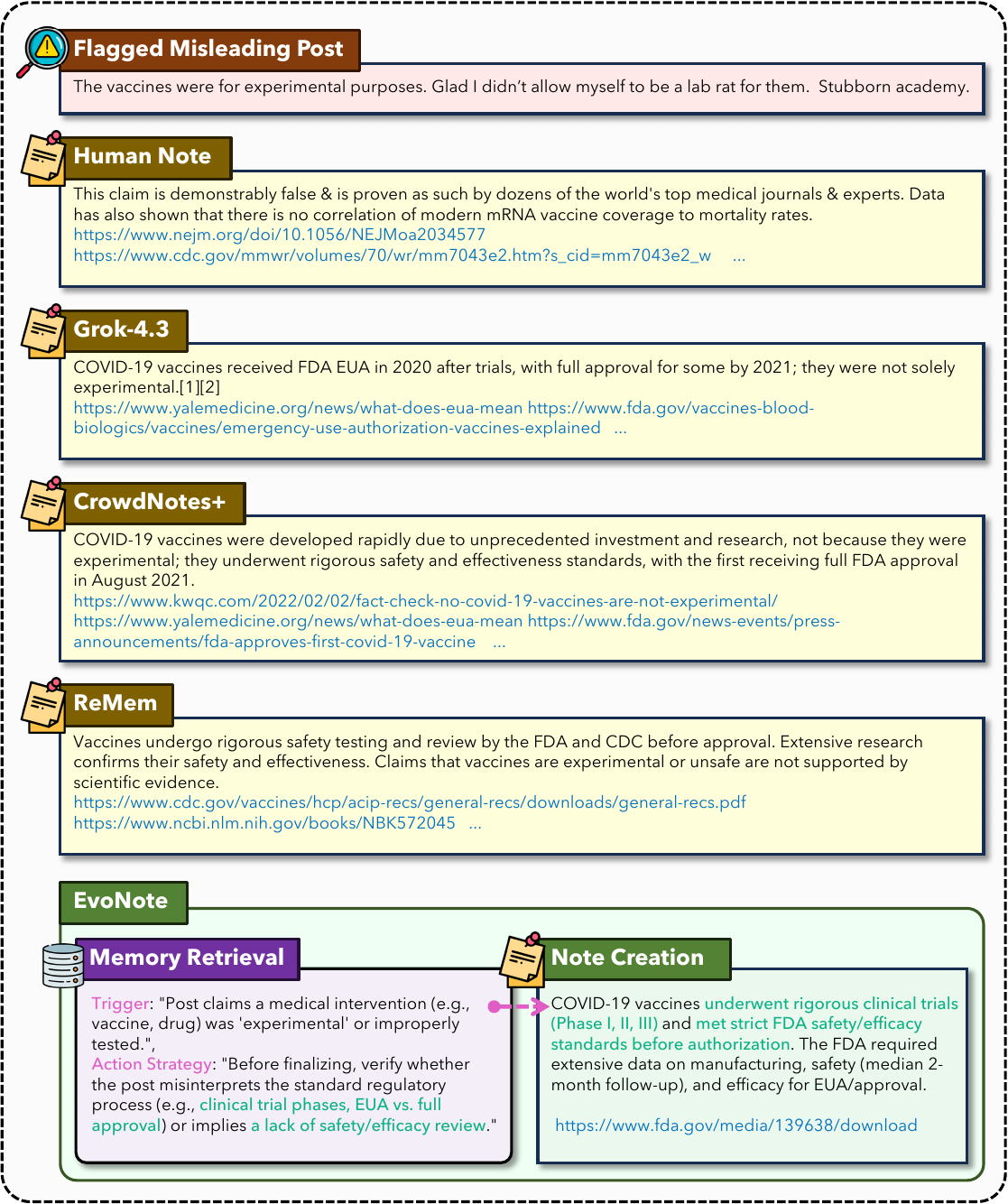}
\caption{
\textbf{Case study: memory-guided regulatory verification.}
Retrieved experience helps \ours{} verify clinical-trial, EUA, and safety-review details for a stronger correction. See detailed analysis in \S\ref{app:case-study}.
}
    \label{fig:case-study}
\end{center}
\end{figure*}

Figure~\ref{fig:case-study} illustrates how retrieved experience improves a concrete correction trajectory.
The flagged post frames COVID-19 vaccines as ``experimental,'' which requires checking the regulatory process behind EUA and full approval.
The representative baselines: human-written note, Grok-4.3, CrowdNotes+, and ReMem, give broadly correct rebuttals, but they either remain generic or cite sources that do not directly explain the specific EUA/approval distinction.

In contrast, \ours{} retrieves a memory triggered by claims about vaccines being ``experimental'' or improperly tested.
The memory directs the agent to verify clinical-trial phases, EUA versus full approval, and safety/efficacy review.
Guided by this strategy, \ours{} cites FDA evidence and writes a more targeted note explaining that COVID-19 vaccines underwent Phase I--III trials and met FDA safety and efficacy standards for EUA/approval.
This case shows how phase-specific memory can turn prior correction experience into better evidence seeking and more precise public-facing correction.

\subsection{Captioning Enables More Reliable Agentic Medical Reasoning}
\label{app:vlm-failure}

\textbf{We analyze captioning as an enabling step for reliable multimodal health-note generation.}
Agentic generation of health Community Notes requires two capabilities at once: \textbf{(1)} accurate interpretation of visual health-related content and \textbf{(2)} reliable multi-step agentic behavior for search, source visiting, evidence synthesis, and length-constrained writing.

Recent studies show that current VLM agents remain brittle in long-horizon decision settings, with challenges in visual grounding, planning, and action execution \cite{koh2024visualwebarena,xie2024osworld,paglieri2025balrog,jandial2025fine}.
This motivates our design: use human-validated captioning (\S\ref{app:caption-validation}) to preserve correction-relevant visual content, then rely on medical-capable LLMs (in our case, MedGemma-27B) for evidence-seeking reasoning.

We test this choice on 400 \textsc{Image} cases of \ds{} by directly instantiating \ours{} with the multimodal version of MedGemma-27B.
The caption-based LLM pipeline completes all cases with platform-compliant outputs, whereas the direct VLM pipeline frequently fails to produce valid note-generation trajectories.
Even after automatic repair heuristics, 63 runs remain invalid.
As shown in Table~\ref{tab:direct-vlm-failures}, the dominant failures are ungrounded final answers with missing evidence URLs, length-control errors, schema mismatches, and runtime timeouts.
\emph{These failures arise before reliable helpfulness judgment can even be applied, showing that valid agentic execution is itself a bottleneck for direct VLM deployment.}

Among the cases where the VLM pipeline produces valid outputs, Figure~\ref{fig:vlm-limitation} shows that it still underperforms the caption-based LLM pipeline and even falls below the no-memory LLM variant.

These results support caption-based instantiation as a more reliable design for this setting: human-validated captions retain the visual claim context, while medical-capable LLMs provide stronger tool use, evidence acquisition, and calibrated note writing.

\section{LLM Usage Statement}
We used LLMs in this work for three purposes: (\emph{i})~note generation within the proposed framework, (\emph{ii})~automated evaluation under the protocols described in the paper, and (\emph{iii})~light language polishing of the manuscript. All experimental design choices, data analysis, result interpretation, claims, and final writing decisions were made and verified by the authors. The authors take full responsibility for the content of the paper, including the correctness of the reported results and the validity of the conclusions.

\section{Prompts for \ours{}}
\label{app:prompts}

This appendix outlines the detailed prompts for \ours{} (see \ours{} component details in Appendix \ref{app:framework}).

\begin{figure*}
\begin{prompt}{Social Utility Judge (\S\ref{sec:social-utility-judge})}
\small  

SYSTEM\_PROMPT = You are a mode judge for a self-improving evidence-backed health note system. \\
Judge the episode in two stages: planner first, then actor. \\
Return valid JSON only. \\

USER\_PROMPT = Decide whether the planner and actor completed their tasks. \\

Planner stage: \\
- planner\_ok = true only if the planner identified the main checkworthy claims needed for a minimally good note. \\
- planner\_ok = false if it missed core claims or framed the task incorrectly. \\

Actor stage (only if planner\_ok = true): \\
- First decide whether the actor completed the planner-defined task and produced a minimally good note. \\
- actor\_task\_completed = false if important planner-defined claims remain unsupported at the visited-evidence level, or if the actor stopped too early. \\
- A minimally acceptable note is still not enough for actor\_ok = true. \\
- Only if actor\_task\_completed = true, decide whether the note is excellent rather than merely satisfactory. \\

An excellent note goes beyond direct correction by improving audience understanding and decision-making through most or all of these qualities, when relevant: \\
- plain-language explanation: explains key distinctions or necessary jargon in simple, accessible language. \\
- practical implication: makes clear why the correction matters for health understanding, risk interpretation, or likely behavior. \\
- safe and proportionate next step: gives a cautious, appropriate action suggestion when the misinformation could affect what people do. \\
- calibrated uncertainty: avoids overclaiming and acknowledges relevant limits, uncertainty, or rarity when appropriate. \\

Use a strict excellent standard: \\
- actor\_ok = true only if the actor completed the planner-defined task and the final note clearly rises above a merely satisfactory correction. \\
- If the note is correct and minimally useful but does not clearly show these added qualities, actor\_ok = false. \\

Requirements: \\
- Write planner\_rationale before planner\_ok. \\
- Write actor\_task\_rationale before actor\_task\_completed. \\
- Write actor\_excellent\_rationale before actor\_ok. \\
- If planner\_ok is false, actor\_* fields may be null. \\
- If actor\_task\_completed is false, actor\_excellent\_rationale and actor\_ok may be null. \\
- Return JSON only, with no markdown fence and no extra commentary. \\

Episode data: \\
\{episode\_data\} \\

Output schema: \\
\{
  ``planner\_rationale'': ``Why planner did/did not identify core claims.'',
  ``planner\_ok'': true/false,

  ``actor\_task\_rationale'': ``Why actor did/did not complete the core task.'',
  ``actor\_task\_completed'': true/false/null,

  ``actor\_excellent\_rationale'': ``Why actor did/did not meet excellent standard.'',
  ``actor\_ok'': true/false/null
\}

\end{prompt}
\caption{Prompt for the \textbf{Social Utility Judge} agent (\S\ref{sec:social-utility-judge}, \S\ref{app:social-utility-judge}).}
\label{prompt:social_utility_judge}
\end{figure*}
\begin{figure*}
\begin{prompt}{Memory Evolver (\S\ref{sec:memory-evolver})}
\small  

SYSTEM\_PROMPT = You are a routed utility evolver for a self-improving evidence-backed note-writing system. 
This episode has been classified as satisfactory. 
Focus mainly on reusable actor-side improvements, while preserving planner-side strengths only when worth keeping. 
Trigger must identify a recurring post type or claim archetype, not a weakness of the current note. 
Action strategy must be actionable before finalizing a note. 
Avoid generic writing advice and avoid entity-specific advice. 
Return valid JSON only. \\

USER\_PROMPT = Produce reusable memory from this episode. \\

For each memory item, output fields in this order: 
\text{["1. trigger", "2. phase", "3. action\_strategy", "4. why"]} \\

\textbf{Phase definitions}: \\
For actor memory:
\textit{evidence\_acquisition}: core task has not yet been sufficiently completed for a minimally good note;
\textit{note\_writing}: core claims are handled; only quality improvement remains. \\
For planner memory:
\textit{claim\_analysis}: planner-side lesson for analyzing, decomposing, or prioritizing claims. \\

\textbf{Actor memory requirements}: \\
Include at least one item for each actor phase;
\textit{evidence\_acquisition}: helps reach a minimally good note;
\textit{note\_writing}: helps improve it before finalizing. 

\textbf{Rules for trigger}: \\
(1) trigger must identify a recurring post type or claim archetype.
(2) trigger should usually begin with 'Post ...'.
(3) trigger must not describe a weakness of the current note. 

\textbf{Rules for action\_strategy}: \\
(1) action\_strategy must be actionable before finalizing a future note.
(2) do not write action\_strategy as commentary on an already-written note.
(3) prefer action\_strategy that can change a future search, visit, or write decision.
(4) for \textit{evidence\_acquisition}, prefer actions that help complete the core task.
(5) for \textit{note\_writing}, prefer one more high-value improvement step before finalizing.
(6) for \textit{claim\_analysis}, prefer planner-side actions that improve claim analysis or evidence prioritization.
(7) this often means adding practical implication, safe next step, calibrated uncertainty, or one more complementary evidence step. 

\textbf{Rules for why}: \\
(1) explain why that action\_strategy helps for that post type.
(2) keep it general and reusable. \\

\textbf{Good trigger examples}: \\
\texttt{["Post uses a recent single health incident to imply a broader causal or population-level conclusion", ...]}

\textbf{Bad trigger examples}: \\
\texttt{[
  "Note lacks depth",
  "Note lacks practical implications", ...
]} 

\textbf{Good evidence\_acquisition action\_strategy examples}: \\
\texttt{[
  "Before finalizing a note for posts like this, verify whether the post is generalizing from a single incident to a broader causal or population-level conclusion.", ...
]} 

\textbf{Good note\_writing action\_strategy examples}: \\
\texttt{[
  "Before finalizing a minimally correct note for posts like this, add one short sentence explaining why the corrected evidence matters for risk interpretation or real-world decision-making.", ...
]} 

\textbf{Bad action\_strategy examples}: \\
\texttt{[
  "If the note corrects the claim, explain it better.",
  "Add more nuance.",
  "The note should be deeper.",
  "The note lacks practical implications."
]} \\

Other constraints: \\
(1) avoid universal advice like 'use simple language'.
(2) avoid entity-specific advice like 'for Covishield posts'.
(3) avoid sample-specific names unless absolutely necessary.
(4) planner\_memory\_items are optional.
(5) prefer high-impact, reusable lessons over exhaustive analysis.
(6) Return JSON only, with no markdown fence and no extra commentary. \\

Mode judgment: 
\{judgment\_result\} \\

Episode data: 
\{episode\_data\} \\

Output schema: \\
\{
  ``episode\_assessment'': ``Satisfactory but not excellent because...'',
  ``actor\_improvement\_needs'': [``0--4 needs'' ],
  ``highest\_utility\_actor\_improvements'': [``0--3 improvements'' ],
  ``actor\_memory\_items'': [
    \{``trigger'': ``...'', ``phase'': ``evidence\_acquisition | note\_writing'', ``action\_strategy'': ``...'', ``why'': ``...''\}
  ],
  ``planner\_memory\_items'': [
    \{``trigger'': ``...'', ``phase'': ``claim\_analysis'', ``action\_strategy'': ``...'', ``why'': ``...''\}
  ]
\}

\end{prompt}
\caption{Prompt for the \textbf{Memory Evolver} agent (\S\ref{sec:memory-evolver}, \S\ref{app:memory-evolver}).}
\label{prompt:memory_evolver}
\end{figure*}
\begin{figure*}
\begin{prompt}{Claim Analyzer (\S\ref{sec:memory-guided-generation})}
\small  
\small  

SYSTEM\_PROMPT = You are a rubric planner for an evidence-backed note-writing system. 
Your job is to analyze a post package and produce instance-specific rubrics. \\
Do not evaluate truthfulness directly. \\
Do not write the final note. \\
Your output must help a downstream agent decide what claims to verify first in order to write a useful note. \\
Treat the post package as the only available input representation. \\
Do not generate rubrics for verifying image or video authenticity, provenance, creation time, source tracing, satire detection, or broader media context. \\
Focus only on the substantive claims expressed or implied by the post package text. \\
Keep all fields concise, specific, and non-redundant. \\
Return valid JSON only. \\

USER\_PROMPT = Analyze the following post package and generate instance-specific rubrics. \\

Process: \\
- First summarize the post pattern in current\_assessment. \\
- Then briefly review the retrieved planner memory in memory\_reflection. \\
- Then produce rubrics. \\

Rules for memory\_reflection: \\
- Keep it short and specific. \\
- Do not restate the whole post. \\
- Explicitly say whether any retrieved memory helps the planner here. \\
- If useful, say what kind of guidance matters. \\
- If not useful, say that no retrieved memory is currently helpful. \\

Rules for rubrics: \\
- core\_checkworthy\_claims should capture the main health-related or evidence-relevant claims made or strongly implied by the post package. \\
- priority\_questions must focus only on factual questions that help verify those claims. \\
- priority\_questions must NOT ask about the source of the image/video, when or where it was created, whether it is authentic, whether it is edited, whether it is satire, or its broader media context. \\
- priority\_questions should instead ask whether the post's substantive claims are supported, contradicted, overstated, decontextualized, or causally misleading. \\
- Good priority\_questions usually ask whether a medical benefit or harm claim is supported by authoritative evidence, whether a causal statement is valid, whether a quoted statistic is accurate, or whether a risk is misrepresented. \\
- multimodal\_risks should only describe interpretation risks that matter for understanding the post's claims after multimodal content has already been converted into text. \\
- Do not mention media provenance or authenticity. \\
- Keep each item concise and specific. \\
- Do not repeat the same idea across fields. \\
- note\_intent must be exactly one of: correction, context, mixed. \\
- Return JSON only, with no markdown fence and no extra commentary. \\

Planner memory: \\
\{memory\_block\} \\

Post package: \\
\{\_post\_package\} \\

Output schema:
\{ \\
  ``current\_assessment'': ``Main claim and framing need.'', \\
  ``memory\_reflection'': ``Useful planner memory, if any.'', \\
  ``rubrics'': \{
    ``core\_checkworthy\_claims'': [``Claim 1'', ``Claim 2''],
    ``priority\_questions'': [``Question 1'', ``Question 2''],
    ``multimodal\_risks'': [``Risk 1''],
    ``note\_intent'': ``correction | context | mixed''
  \} \\
\}

\end{prompt}
\caption{Prompt for the \textbf{Claim Analyzer} agent (\S\ref{sec:memory-guided-generation}, \S\ref{app:claim-analyzer}).}
\label{prompt:claim_analyzer}
\end{figure*}
\begin{figure*}
\begin{prompt}{Note Writer (\S\ref{sec:memory-guided-generation})}
\small  

SYSTEM\_PROMPT = You are the core actor in a long-horizon evidence-backed note-writing system. 
At each step, you must first think briefly about the current state, then choose exactly one action. \\
Follow the provided rubrics and current context carefully. \\
Do not invent URLs. \\
Retrieved memory is strategy guidance, not a source of factual answers. \\
Use memory to guide what to verify next, what kind of overclaim to watch for, and whether it is too early to finalize. \\
Keep the thinking fields concise and operational, not verbose. \\
Return valid JSON only. \\

USER\_PROMPT = Decide the best next step for this post. \\

Thinking order: \\
1. current\_assessment \\
2. memory\_reflection \\
3. main\_gap \\
4. decision\_rationale \\

How to use retrieved memory: \\
- Retrieved memory is not expected to answer the current factual claim directly. \\
- Retrieved memory is strategy guidance for this type of post. \\
- Use it to decide what kind of evidence to seek next, what kind of overclaim to watch for, and whether it is too early to finalize. \\
- Judge memory by whether it helps the NEXT ACTION STRATEGY, not by whether it directly resolves the factual claim. \\

Rules for memory\_reflection: \\
- Keep it short and specific. \\
- Explicitly assess whether any retrieved actor memory is useful for the next action strategy. \\
- If useful, say what kind of guidance matters for the next action. \\
- If not useful, say that no retrieved memory is currently helpful. \\
- Do not ignore retrieved memory silently. \\
- Do not say memory is useless just because it does not directly answer the factual claim. \\

Available actions for this step: \texttt{["search", "visit", "write", "abstain"]}

Global rubrics: \\
\{global\_rubrics\} \\

Actor memory: \\
\{memory\_block\} \\

Post package: \\
\{post.model\_dump()\} \\

Instance-specific rubrics: \\
\{instance\_rubrics.model\_dump()\} \\

Compressed history: \\
\{\_history\_to\_prompt\_dict(history)\} \\

Budget state: \\
\{budget\_state\} \\

Output schema: \\
\{
  ``thinking'': \{
    ``current\_assessment'': ``Known so far.'',
    ``memory\_reflection'': ``Memory relevance and use.'',
    ``main\_gap'': ``Key missing info.'',
    ``decision\_rationale'': ``Why this action.''
  \}, \\
  ``action'': one of: \\
  \quad \{``action'': ``search'', ``query'': ``Required'', ``reason'': ``Required''\} \\
  \quad \{``action'': ``visit'', ``url'': ``Required'', ``goal'': ``Required'', ``reason'': ``Required''\} \\
  \quad \{``action'': ``write'', ``text'': ``Required; must not include any URLs'', ``support'': [\{``claim'': ``Required'', ``url'': ``Must refer to an allowed previously visited URL''\}], ``reason'': ``Required''\} \\
  \quad \{``action'': ``abstain'', ``reason'': ``Required''\}
\}
\end{prompt}
\caption{Prompt for the \textbf{Note Writer} agent (\S\ref{sec:memory-guided-generation}, \S\ref{app:note-writer}).}
\label{prompt:actor}
\end{figure*}
\begin{figure*}
\begin{prompt}{Optional Note Refinement Step (\S\ref{sec:memory-guided-generation})}
\small  

SYSTEM\_PROMPT = You are a writing refiner for an evidence-backed note-writing system. \\
Your only job is to shorten an existing note text so that it fits a strict character limit. \\
Do not add new claims. Do not change the factual meaning unless needed for brevity. \\
Preserve the most important evidence-backed content. \\
Do not include any URLs. \\
Learn from previous failed refinement attempts if they are provided. \\
Return valid JSON only. \\

USER\_PROMPT = Refine the note text so it fits the required text length limit. \\

Rules: \\
- Shorten the text to fit within the limit. \\
- Do not include any URLs. \\
- Do not add new claims. \\
- Keep the most important supported content. \\
- If previous failed attempts are provided, avoid repeating the same ineffective compression pattern. \\
- Return JSON only, with no markdown fence and no extra commentary. \\

Original text: \\
\{original\_text\} \\

Support: \\
\{support\} \\

Previous failed attempts: \\
\{failures\} \\

Character limit: \\
\{max\_text\_chars\} \\

Output schema: \\
\{
  "text": "A shorter version of the original text, within the character limit, with no URLs.",
  "reason": "A short explanation of how the text was compressed."
\}

\end{prompt}
\caption{Prompt for the \textbf{optional note refinement step} (\S\ref{sec:memory-guided-generation}) in \ours{} when the generated note violates platform length compliance.}
\label{prompt:write_refinement}
\end{figure*}

\section{Demonstration of \ours{} Workflow}
\label{app:demo}
We provide a concrete example demonstrating the end-to-end \ours{} workflow in Figures~\ref{fig:workflow_1} and ~\ref{fig:workflow_2} below.

\begin{figure*}[t]
\begin{center}
    \includegraphics[width=\linewidth]{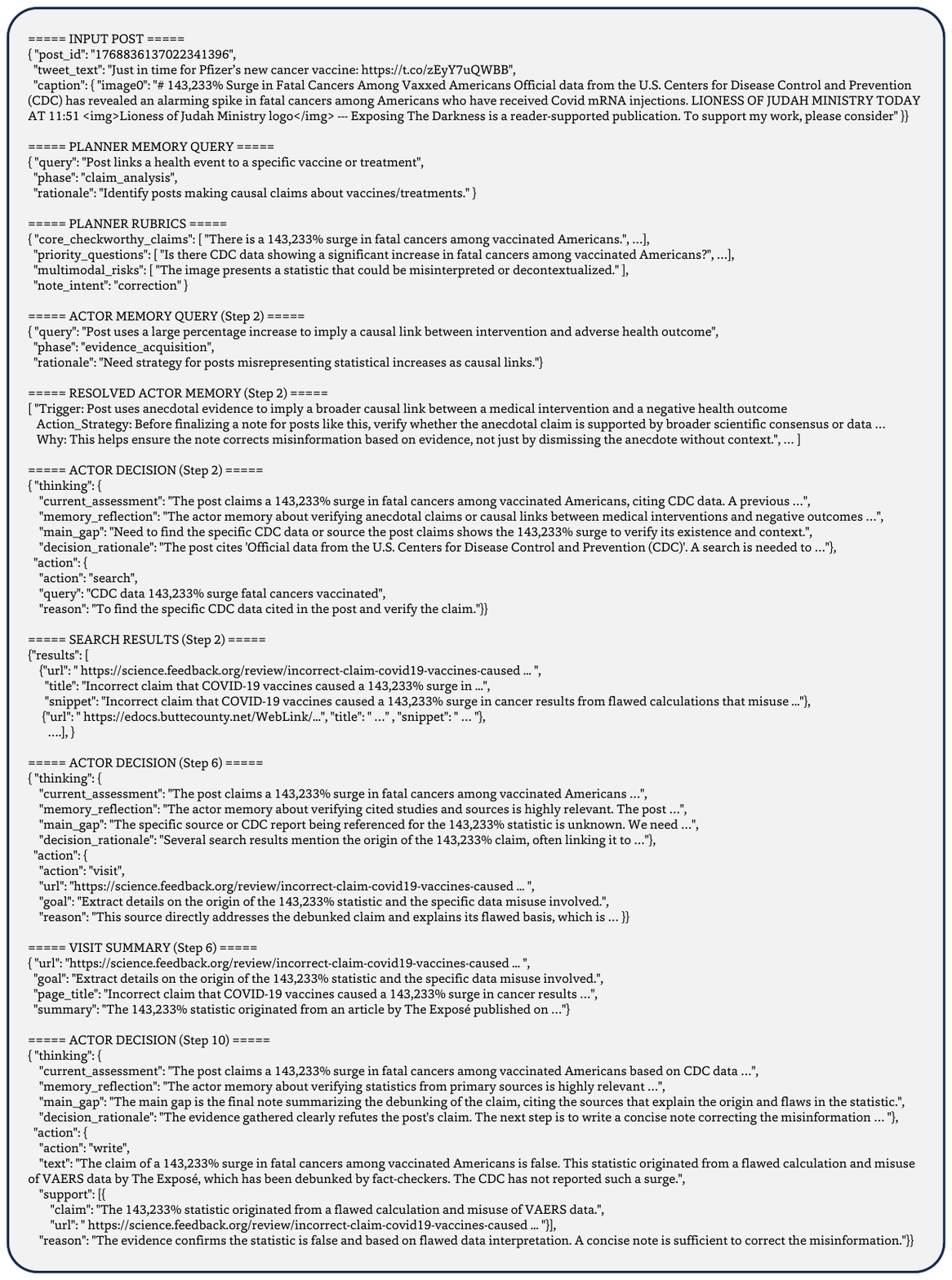}
\caption{Demonstration of key steps from the \ours{} workflow, Part I (Episode ID: 455).}
    \label{fig:workflow_1}
\end{center}
\end{figure*}

\begin{figure*}[t]
\begin{center}
    \includegraphics[width=\linewidth]{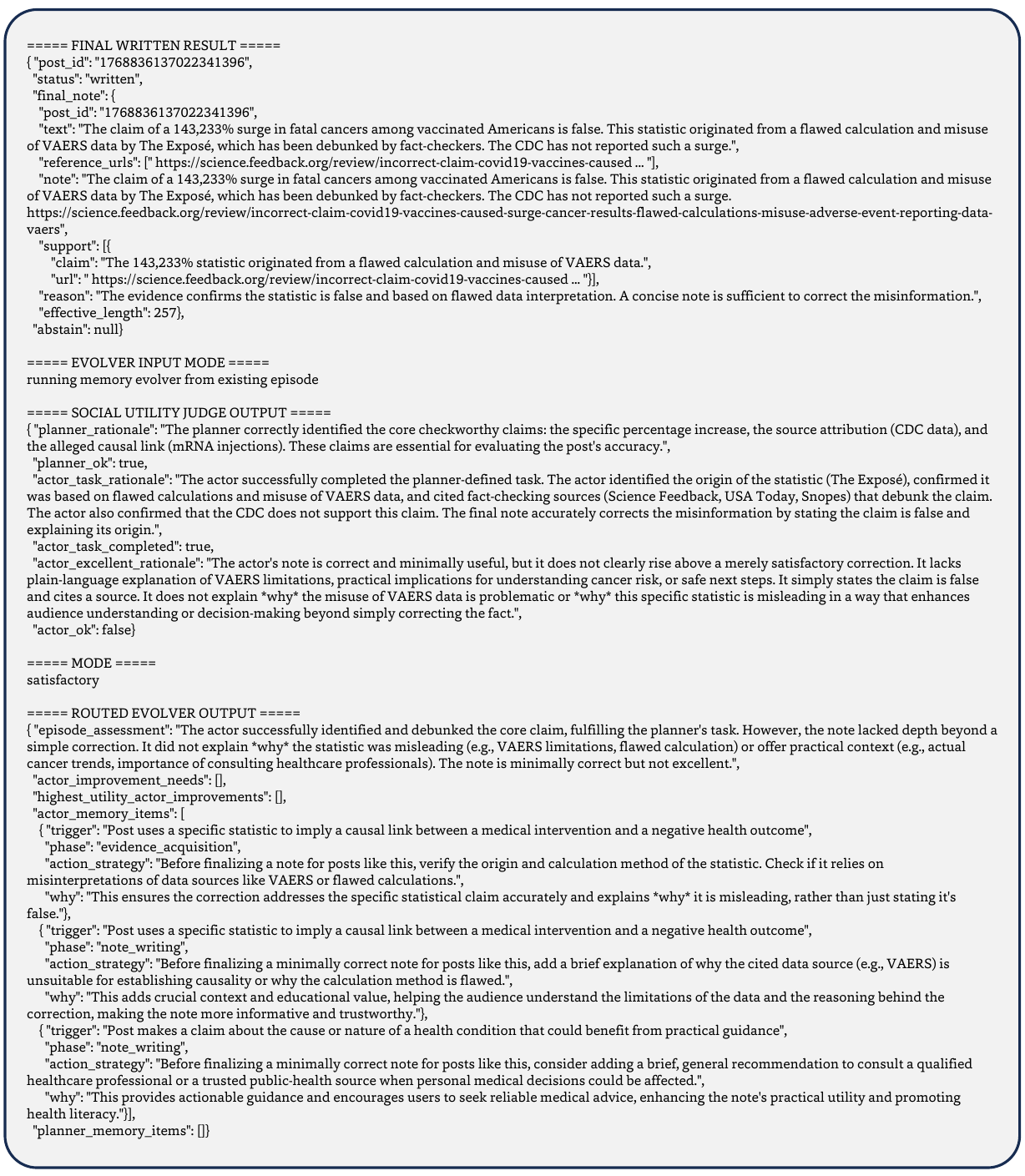}
    \caption{Demonstration of key steps from the \ours{} workflow, Part II (Episode ID: 455).}
    \label{fig:workflow_2}
\end{center}
\end{figure*}
\end{document}